\UseRawInputEncoding
\documentclass[10pt,twocolumn,letterpaper]{article}

\usepackage[pagenumbers]{cvpr} % To force page numbers, e.g. for an arXiv version

\usepackage{pifont}
\newcommand{\cmark}{\ding{51}}
\newcommand{\xmark}{\ding{55}}
\newcommand{\lgcmark}{\textcolor{green}{\cmark}}
\newcommand{\lgxmark}{\textcolor{red}{\xmark}}
\usepackage{graphicx}
\usepackage{caption} % for \captionof

\newcommand{\vebenc}{\textsc{VEBench}\xspace}

\usepackage{cuted}        % widetext (full-width block in 2-col)
\usepackage{tcolorbox}
\tcbuselibrary{listings,breakable}
\usepackage{listings,xcolor}
\usepackage{capt-of}      % for \captionof outside floats

\lstdefinestyle{promptstyle}{
  basicstyle=\ttfamily\scriptsize,
  breaklines=true,
  columns=fullflexible,
  backgroundcolor=\color{gray!4},
  frame=single, rulecolor=\color{gray!40},
  showstringspaces=false, keepspaces=true,
}

\definecolor{cvprblue}{rgb}{0.21,0.49,0.74}
\usepackage[pagebackref,breaklinks,colorlinks,allcolors=cvprblue]{hyperref}
\usepackage{multirow}
\usepackage[table]{xcolor}
\usepackage{float}
\definecolor{MyDarkRed}{rgb}{0.8,0.02,0.02}

\def\confName{CVPR}
\def\confYear{2026}

\title{\vebenc: Benchmarking Large Multimodal Models for \\Real-World Video Editing}

\author{
Andong Deng$^{1,2,\ddagger}$,
Dawei Du$^{1}$,
Zhenfang Chen$^{1}$,
Wen Zhong$^{1}$,
Fan Chen$^{1}$,
Guang Chen$^{1}$, \\
Chia-Wen Kuo$^{1}$,
Longyin Wen$^{1}$,
Chen Chen$^{2}$,
Sijie Zhu$^{1,*}$\\
$^{1}$ByteDance Intelligent Creation \quad
$^{2}$CRCV, University of Central Florida
}

\begin{document}

\twocolumn[{%  <--- NOTE: "[" followed immediately by "{"
\maketitle

\vspace{-3em}
\begin{center}
  \includegraphics[width=\linewidth]{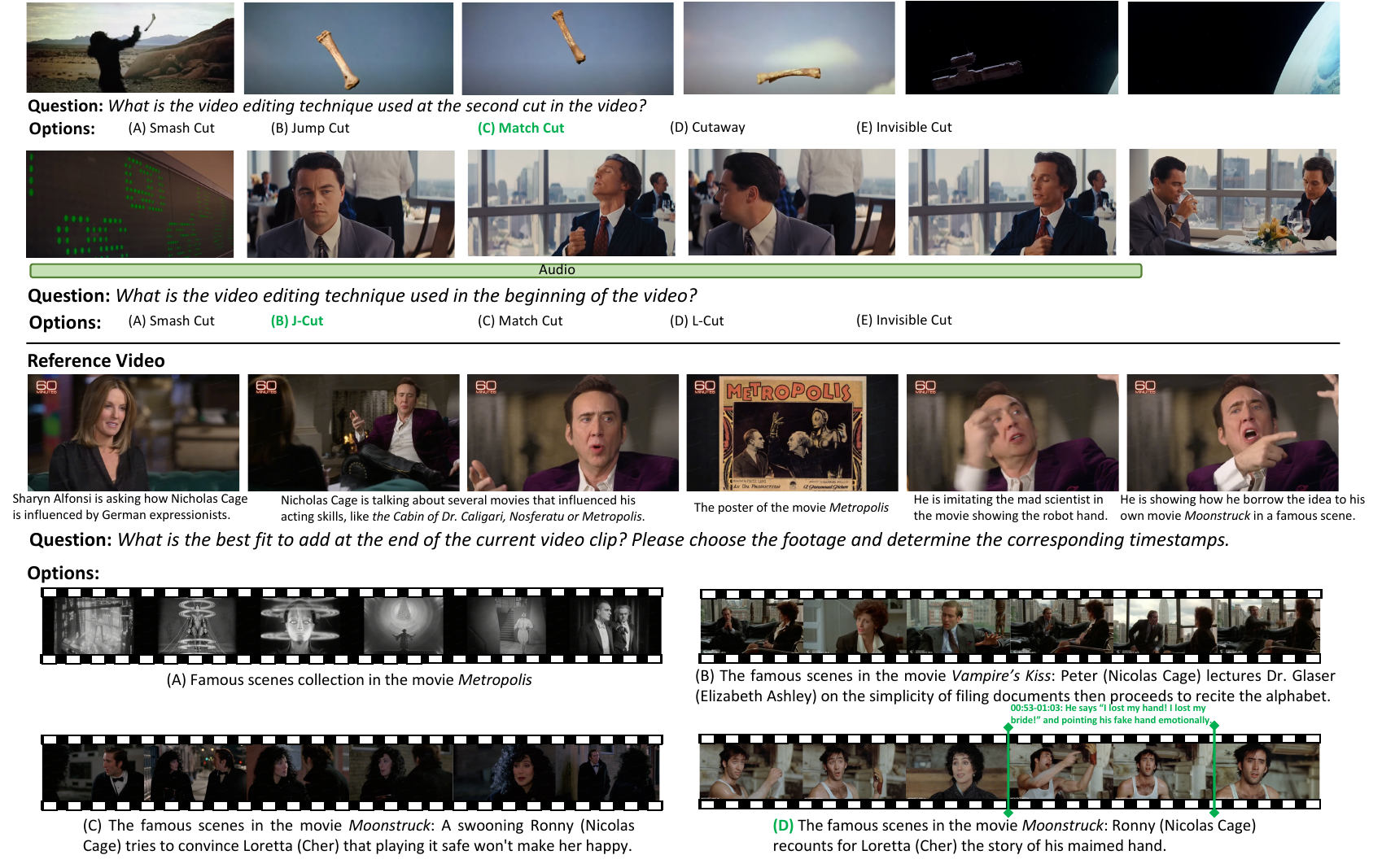}
  \vspace{-2em}  % small extra tweak between image and caption
  \captionsetup{type=figure}
  \captionof{figure}{
    Examples of the two tasks in \textbf{\vebenc}.
    \textbf{Video Editing Technique Recognition} (top) detects the editing technique at a specified position;
    \textbf{Video Editing Operation Simulation} (bottom) first selects the most suitable footage from a candidate list and then localizes the proper segment based on a reference video.
    Correct answers are indicated in green.
  }
  \label{fig:teaser}
\end{center}
\vspace{0.8em}
}] %  <--- 

\let\thefootnote\relax\footnotetext{$*$ Corresponding author, sijiezhu@bytedance.com}
\let\thefootnote\relax\footnotetext{$\ddagger$ Work was done during the internship at ByteDance, San Jose, USA}

\begin{abstract}
Real-world video editing demands not only expert knowledge of cinematic techniques but also multimodal reasoning to select, align, and combine footage into coherent narratives.
While recent Large Multimodal Models (LMMs) have shown remarkable progress in general video understanding, their abilities in multi-video reasoning and operational editing workflows remain largely unexplored.
We introduce \textbf{\vebenc}, the first comprehensive benchmark designed to evaluate both editing knowledge understanding and operational reasoning in realistic video editing scenarios.
\vebenc contains $3.9K$ high-quality edited videos (over $257$ hours) and $3,080$ human-verified QA pairs, built through a three-round human–AI collaborative annotation pipeline that ensures precise temporal labeling and semantic consistency.
It features two complementary QA tasks: 1) \textbf{Video Editing Technique Recognition}, assessing models' ability to identify $7$ editing techniques using multimodal cues; and 2) \textbf{Video Editing Operation Simulation}, modeling real-world editing workflows by requiring the selection and temporal localization of relevant clips from multiple candidates.
Extensive experiments across proprietary (\eg, Gemini-2.5-Pro) and open-source LMMs reveal a large gap between current model performance and human-level editing cognition.
These results highlight the urgent need for bridging video understanding with creative operational reasoning.
We envision \vebenc as a foundation for advancing intelligent video editing systems and driving future research on complex reasoning. Project available at: \href{https://vebench.github.io/}{\textcolor{magenta}{https://vebench.github.io/}}.
%~\zf{need improvement to show the value of the task, the effort of the dataset and the finding/conclusion of our paper.}
\end{abstract}
    
\section{Introduction}
\label{sec:intro}

% Keypoint 1: Video editing itself

% (1) Background. Video editing = knowledge + operation.

% (2) Existing benchmarks. They focus solely on understanding (knowledge), we propose our own knowledge-based QA and further propose operation-based QA to simulate real-world video editing workflow.

% Keypoint 2: Video editing as a complex video reasoning task (multi-video multimodal reasoning).

% (1) Background. Video reasoning 

% (2) Existing video reasoning benchmarks. Limited in single video?

% paragraph 1: the background of video understanding and video reasoning. And the role of video editing in these two field.

% paragraph 2: existing benchmarks that related to video editing and filmmaking. a. VEU-Bench (CVPR 2025) b. ShotBench (https://vchitect.github.io/ShotBench-project/).

% paragraph 3: brief introduction to our \vebenc. The difference, the advantages. And we introduce video reasoning in multiple videos, which is not considered by existing videos benchmarks.

% paragraph 4: the annotation process. and statistics. baseline model based on QWen-2.5 Omni

% final paragraph is the itemized contributions. 

Understanding and reasoning over videos~\cite{Maaz2023VideoChatGPT,qwen3vl,google2025gemini25pro} are fundamental challenges in computer vision and multimodal intelligence. 
Traditional video understanding focuses on recognizing actions~\cite{kay2017kinetics,soomro2012ucf101}, events~\cite{7298698,zhou2018towards}, and objects~\cite{xu2018youtube,Perazzi2016} within a single video, while recent work extends toward temporal~\cite{fu2024videomme,wu2024longvideobench,qi2025vcr}, causal~\cite{cheng2025video,Deng_2025_CVPR,li2022from}, and cross-modal~\cite{li2025omnivideobench} reasoning.
In contrast, video editing~\cite{dancyger2018technique} presents a far more demanding setting that requires both expert-level editing knowledge and the ability to manipulate and combine video footage into coherent narratives.
Situated at the intersection of video understanding and reasoning, editing serves as a real-world testbed for studying complex multimodal cognition—highly relevant to filmmaking~\cite{frierson2018film,metz1991film}, short-form content creation~\cite{wang2021research,huber2019b}, and professional media production~\cite{shyles2007art,johnson2016shoot}.
% ~\zf{need some references here to support your claims}

Several recent benchmarks have begun to explore domains tangentially related to video editing and filmmaking~\cite{li2025veu,liu2025shotbench,wang2025cinetechbench,gu2024edit3k,Xu_2025_CVPR}.
VEU-Bench~\cite{li2025veu} investigates a range of editing techniques such as cuts and transitions, formulating questions that span from recognition to reasoning.
ShotBench~\cite{liu2025shotbench} emphasizes shot-level analysis and cinematic attributes, including composition and camera movement.
While these benchmarks have made valuable contributions toward advancing video understanding in editing-related contexts, they remain limited in two crucial aspects.
First, they primarily evaluate \textit{knowledge-based comprehension} on very short clips (e.g., the average duration of VEU-Bench~\cite{li2025veu} is only $6.22$s), relying almost exclusively on visual cues without considering audio–visual synchronization or long-range temporal context, both of which are indispensable for real-world video editing.
Second, they overlook the operational dimension of editing: the process of selecting, aligning, and combining footage from multiple videos to create coherent narratives.
This dimension requires multi-video reasoning and intent-aware decision-making, which remain largely absent in current LMMs.
% ~\zf{saying only one difference is risky. Could you please add at least one more difference to "edit tech recognition?"} 
% and (2) they are restricted to understanding one short video, leaving open the challenge of \textit{multi-video reasoning} that is central to actual editing scenarios.

To bridge these gaps, we introduce \textbf{\vebenc}, the first benchmark that unifies both editing \textit{knowledge understanding} and \textit{operational reasoning} in a single framework.
As illustrated in Figure~\ref{fig:teaser}, \vebenc defines two complementary QA paradigms that together approximate human editing cognition:
\begin{enumerate}
    \item \textbf{Video Editing Technique Recognition} evaluates a model's ability to perceive and interpret editing techniques using both visual and auditory cues in long-form videos.
    Unlike prior datasets restricted to short clips, \vebenc features videos averaging $213$ seconds with $3.27$ annotated editing cuts per clip, substantially increasing contextual depth and annotation difficulty.

    \item \textbf{Video Editing Operation Simulation} extends beyond recognition to emulate real-world editing workflows.
    Given a reference video, the model must select and temporally ground the most relevant segments from multiple candidate videos, mirroring the decision-making process of professional editors.
    This task introduces the novel dimension of \textbf{multi-video reasoning}, assessing whether models can infer editing intentions and construct narrative continuity across independent sources.
\end{enumerate}
Together, these two tasks decompose the complex process of human editing into perception-level understanding and operation-level reasoning, allowing \vebenc to systematically evaluate the progression of multimodal models from passive video comprehension toward active, human-like video editing intelligence.

% ~\zf{these numbers are exciting and do not show our effort. Could we use how many edit techniques are covered, how many annotation hours do we spend?}
% ~\zf{gap to what? human performance? it seems that we do not have human study}
Beyond task formulation, the construction of \vebenc itself posed substantial challenges due to its multi-video, multimodal, and reasoning-intensive nature.
We construct \vebenc through a rigorous annotation process involving expert-curated video editing scenarios, ensuring high-quality and diverse coverage of both editing techniques and operational workflows. 
The benchmark comprises $3.9K$ high-quality editing videos and $3,080$ QA pairs, covering a broad spectrum of cinematic and vlog-style content.
All annotations are produced through a three-round human–AI collaborative pipeline, involving over $1,400$ working hours by a professional annotation team.
This process ensures fine-grained temporal accuracy, multimodal synchronization, and semantic consistency across editing scenarios.
Extensive evaluations of both proprietary (e.g., Gemini-2.5-Pro~\cite{google2025gemini25pro}) and open-source (e.g., Qwen3-VL~\cite{qwen3vl}) LMMs reveal that even state-of-the-art systems struggle on both subtasks. It highlights the substantial gap between current model capabilities and human-level video editing reasoning.
Our contributions can be summarized as follows:
\begin{itemize}
    \item We formulate a new, human-inspired video editing task that unifies \textit{knowledge understanding} and \textit{operational reasoning}, providing a practical testbed for evaluating complex multimodal and multi-video reasoning.
    \item We introduce \textbf{\vebenc}, a professionally annotated benchmark featuring diverse, fine-grained editing scenarios collected from real-world videos and curated through a three-round human–AI collaborative process.
    \item We present comprehensive evaluations across leading proprietary and open-source LMMs, showing that even the strongest models perform far below human-level on both subtasks, underscoring the benchmark's difficulty and potential to drive future progress in video reasoning and editing intelligence.

    % \item We present extensive annotations, statistics, and baselines, including a strong Qwen2.5-Omni model, establishing \vebenc as a challenging benchmark for future research in Video-LLMs.
    % ~\zf{the summary is not concise and powerful. Maybe we can summarize it into three aspects, task, dataset and evaluation. 1). a new task that is both practical and a good testbed for complex reasoning; 2). a dataset that is finegrained annotated, labor-intensive and valuable; 3). extensive experiment evaluation showing the limitation of current SOTA models and the value of our new task and dataset.}
\end{itemize}

% \begin{figure*}
%   \centering
%   \begin{subfigure}{0.68\linewidth}
%     \fbox{\rule{0pt}{2in} \rule{.9\linewidth}{0pt}}
%     \caption{An example of a subfigure.}
%     \label{fig:short-a}
%   \end{subfigure}
%   \hfill
%   \begin{subfigure}{0.28\linewidth}
%     \fbox{\rule{0pt}{2in} \rule{.9\linewidth}{0pt}}
%     \caption{Another example of a subfigure.}
%     \label{fig:short-b}
%   \end{subfigure}
%   \caption{Example of a short caption, which should be centered.}
%   \label{fig:short}
% \end{figure*}

\section{Related Work}
\label{sec:formatting}

%-------------------------------------------------------------------------
\noindent\textbf{Video Reasoning of Video LLMs.}
Recent advances in LMMs have substantially improved temporal perception and reasoning in video understanding~\cite{fu2024videomme,wu2024longvideobench,hu2025videommmu,song2025videommlu,hu2025videommmu,deng2025scivideobench}.
VideoChatGPT~\cite{Maaz2023VideoChatGPT} employs a simple, straightforward spatio-temporal pooling mechanism for visual representation, while the LLaVA series~\cite{liu2023visual,li2024llavanext} introduces the \textit{anyres} technique to unify the processing of both image and video inputs. 
Meanwhile, LongVA~\cite{zhang2024longva} and LongVILA~\cite{chen2024longvila} extend the temporal context window of LLMs through multi-stage training, enabling reasoning over long-form videos that span several minutes. 
The InternVL~\cite{zhu2025internvl3} and Qwen-VL~\cite{qwen3vl} families further enhance multimodal alignment and instruction following abilities across diverse video understanding tasks.
Despite these advancements, the performance of current LMMs on video editing remains underexplored. However, video editing is a highly demanding real-world application that requires not only an understanding of visual–narrative coherence but also sophisticated multimodal reasoning across multiple video segments. To bridge this gap, we introduce \vebenc, a domain-specific benchmark that rigorously evaluates LMMs in real-world editing scenarios, encompassing both expert editing knowledge and practical editing workflows.

% It demands not only understanding of visual-narrative coherence, but also mastery of the operational logic behind editing decisions, such as shot selection, transition design, and temporal continuity.~\zf{need some modification}

%-------------------------------------------------------------------------
\noindent\textbf{Video Editing Benchmarks.}
Video editing plays a crucial role in filmmaking, vlogging, and content creation, requiring both aesthetic judgment and narrative reasoning. 
However, most existing video editing benchmarks mainly emphasize recognition-level understanding (e.g., visual effects, shot understanding, and cut-type recognition) rather than operational reasoning.
For example, Edit3K~\cite{gu2024edit3k} focuses on aesthetic evaluation of visual effects within individual video clips. 
ShotBench~\cite{liu2025shotbench} designs video-based QA tasks spanning attributes such as shot types, lighting, and camera movements, forming a comprehensive recognition benchmark. 
EditVid-QA~\cite{Xu_2025_CVPR} collects edited videos from social media to assess model understanding of meme-style edits and visual effects. 
Furthermore, VEU-Bench~\cite{li2025veu} extends this line of work by examining video editing from intra-frame to inter-shot perspectives and establishing a three-level evaluation protocol covering shot, transition, and cut-type classification.
Despite these efforts, current benchmarks remain limited in their ability to capture complex, real-world video editing scenarios. As shown in Table~\ref{tab:vebench_comparison}, they mainly focus on perceptual or recognition-oriented evaluations in short temporal context and ignore multimodal cues like audio to understand editing intent. Moreover, they neglect the operational challenges intrinsic to real-world editing, which requires reasoning across multiple video sources to construct a coherent and purposeful storyline.
To bridge this gap, we introduce \textbf{\vebenc}, the first benchmark that integrates real-world editing operations into the evaluation of multimodal reasoning. Unlike prior datasets, \vebenc\ jointly measures (i) knowledge of editing principles and (ii) reasoning over footage selection, editing intent, and fine-cut design, thereby offering a more comprehensive and realistic testbed for assessing the capabilities of large multimodal models in video editing contexts.

\section{Benchmark}

\begin{table}[ht]
\centering
\setlength{\tabcolsep}{6pt} % adjust spacing
\renewcommand{\arraystretch}{1.2} % adjust vertical spacing
\scalebox{0.6}{
\begin{tabular}{l
>{\centering\arraybackslash}p{1.cm}
>{\centering\arraybackslash}p{1.2cm}
>{\centering\arraybackslash}p{1.4cm}
>{\centering\arraybackslash}p{1.2cm}
>{\centering\arraybackslash}p{1.0cm}
>{\centering\arraybackslash}p{1.0cm}
>{\centering\arraybackslash}p{1.0cm}}
\toprule
\multirow{2}{*}{\textbf{Datasets}} &
% \multirow{2}{*}{\textbf{\#Videos}} &
\multirow{2}{*}{\textbf{\#QAs}} &
\multicolumn{1}{c}{\textbf{Avg. Vid.}} &
\multicolumn{1}{c}{\textbf{Avg. Cuts}} &
\multirow{2}{*}{\textbf{Reasoning}} &
\multicolumn{1}{c}{\textbf{Multi-}} &
\multicolumn{1}{c}{\textbf{Editing}} &
\multirow{2}{*}{\textbf{Audio}} \\
 & & \textbf{Duration} & \textbf{per Vid.} & & \textbf{Video} & \textbf{Workflow} & \\
\midrule
Edit3K~\cite{gu2024edit3k}       & -  & - & - & \lgxmark & \lgxmark & \lgxmark & \lgxmark \\
ShotBench~\cite{liu2025shotbench}  & 3,572 & - & - & \lgxmark & \lgxmark & \lgxmark & \lgxmark \\
EditVid-QA~\cite{Xu_2025_CVPR}  & 439 & 9.2s & - & \lgxmark & \lgxmark & \lgxmark & \lgxmark \\
VEU-Bench~\cite{li2025veu}      & 4,382 & 6.2s & 1.0 & \lgcmark & \lgxmark & \lgxmark & \lgxmark \\ \hline
\textbf{\vebenc}  & 3,080 & 237s & 3.3 & \lgcmark & \lgcmark & \lgcmark & \lgcmark \\
\bottomrule
\end{tabular}}
\caption{Comparisons between \vebenc and other video editing benchmarks. \vebenc integrates long-term videos and multi-video multimodal reasoning to emulate real editing scenarios.}
\label{tab:vebench_comparison}
\vspace{-5pt}
\end{table}

%\dd{I add some motivation for the benchmark.}
Human video editing is a two-stage cognitive process that integrates both knowledge understanding and goal-directed reasoning. 
Professional editors first perceive and interpret editing techniques, such as \textit{Match Cut}, \textit{Jump Cut}, or \textit{L-Cut}, to maintain visual and narrative continuity.
Then they engage in operational reasoning, actively selecting, trimming, and arranging footage from multiple sources to construct coherent and emotionally resonant storylines.
Together, these two stages reflect the essential duality of human editing cognition: understanding how edits work and reasoning why they are applied.
To emulate this dual structure, \vebenc defines two subtasks that together approximate the real-world editing workflow as follows.
\begin{figure}[t]
  \centering
   \includegraphics[width=1\linewidth]{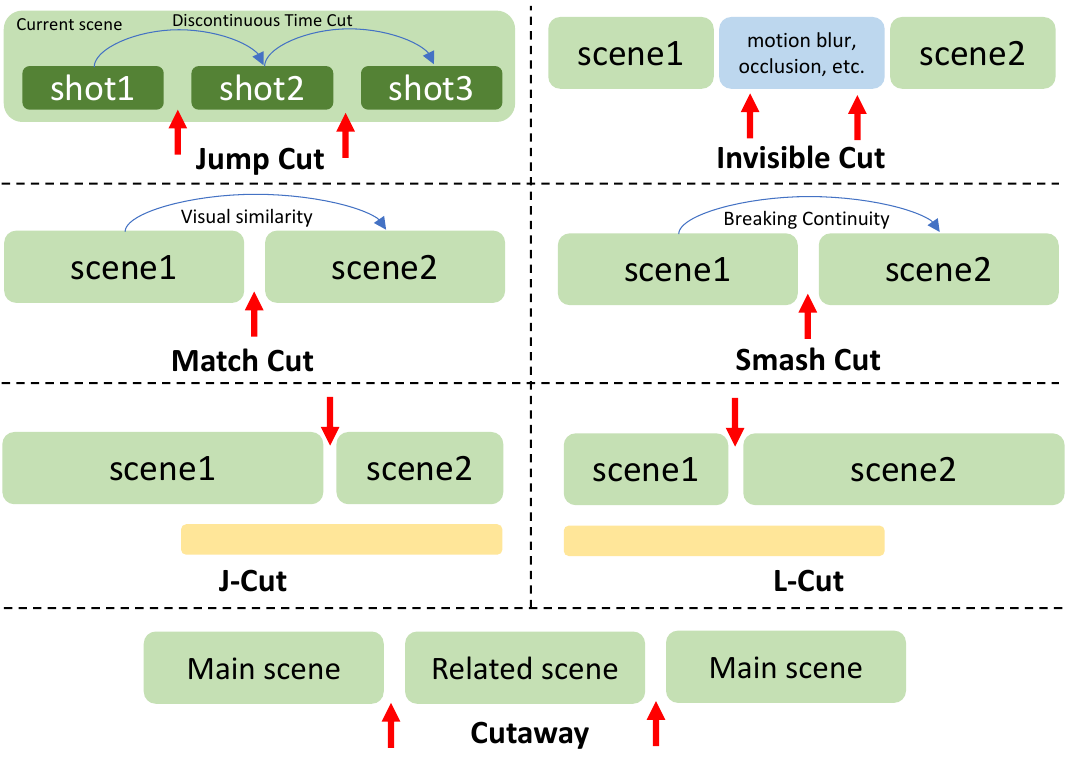}

   \caption{Video Editing Techniques Illustrations. The red arrows indicate the transition in video, and the yellow bars indicate the audio track. Best viewed in color.}
   \label{fig:techs}
   \vspace{-10pt}
\end{figure}

\subsection{Task Definition}
\begin{figure*}[t]
  \centering
   \includegraphics[width=0.95\linewidth]{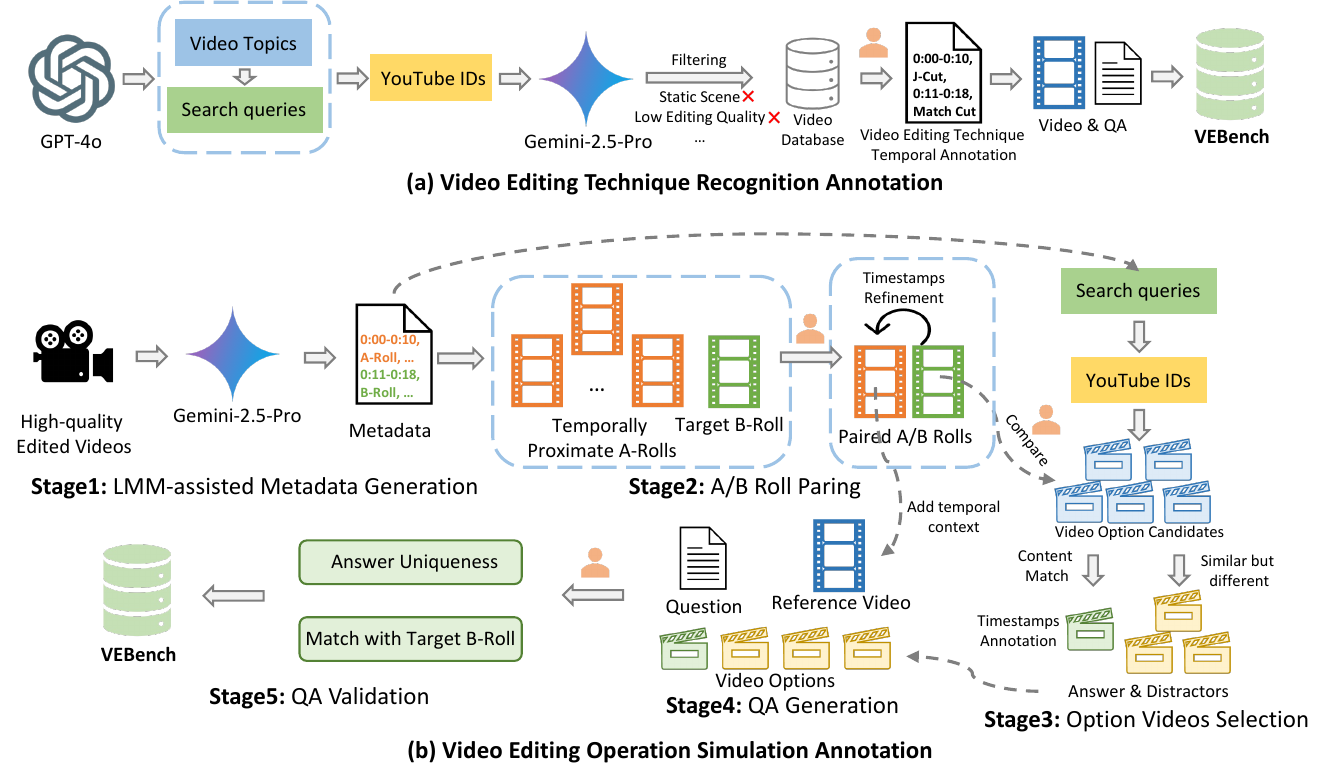}

   \caption{Annotation pipelines of the two \vebenc\ subtasks. (a) \textbf{TechRec}: GPT-4o provides YouTube video search queries and Gemini-2.5-Pro filter video candidates; human annotators inspect the videos and finalize the editing technique labels. (b) \textbf{OpSim}:  Gemini-2.5-Pro analyzes the selected high-quality edited videos and generate metadata; human annotators perform A/B Roll paring and refine timestamps; human annotators select option videos from the retrieved pool and annotate timestamp for the ground truth video; QAs are automatically generated based on the annotated data and human annotators perform the QA validation.}
   \label{fig:annotation}
   \vspace{-5pt}
\end{figure*}

%These techniques require not only comprehensive visual understanding but also precise perception of audio-visual synchronization, a crucial aspect often overlooked in prior video editing benchmarks.
%Specifically, \textit{J-Cut} and \textit{L-Cut} manipulate audio timing to achieve smoother transitions. The former introduces the next scene's audio before the visual cut, while the latter extends the current scene's audio beyond the visual change.
%\textit{Jump Cut} deliberately breaks the temporal continuity within a single scene to convey rhythm, time compression, or psychological tension.
%\textit{Cutaway} inserts a brief contextual or emotional shot before returning to the main sequence.
%\textit{Invisible Cut} hides the transition using motion blur, occlusion, or darkness to create a ``one-take'' illusion.
%\textit{Match Cut} connects two shots with strong graphical, compositional, or thematic resemblance to maintain visual continuity, while \textit{Smash Cut} employs a sudden contrast, either visual or auditory, to emphasize shock or abrupt narrative change. 

The first subtask, \textbf{Video Editing Technique Recognition (TechRec)}, focuses on identifying seven editing techniques, which together form the fundamental constructs of modern filmmaking and vlog production~\cite{dancyger2018technique}.
These techniques span both temporal transitions (e.g., \textit{L-/J-/Jump/Smash Cuts}) and semantic transitions (e.g., \textit{Cutaway, Match Cut, Invisible Cut}), requiring reasoning across visual, auditory, and narrative dimensions (see Figure~\ref{fig:techs}).

Unlike conventional video QA tasks, \textbf{TechRec} demands fine-grained multimodal perception, where success depends on recognizing subtle cues such as cross-scene audio continuity, rhythmic pacing, and visual composition alignment.
For instance, J-/L-Cuts rely on precise audio–visual synchronization, Jump Cuts intentionally disrupt temporal coherence, while Match and Smash Cuts depend on high-level semantic contrast or continuity.
In this task, models must determine the specific editing technique used at each transition point within a finely edited video—requiring advanced temporal reasoning, multimodal synchronization, and contextual understanding.
Definitions and representative examples of all techniques are included in the \textbf{Supp. Sec. 1}.

The second subtask, \textbf{Video Editing Operation Simulation (OpSim)}, mirrors the real-world decision-making process in professional video editing. 
Given a fixed topic, human editors typically select suitable footage from a large pool and organize a coherent storyline by placing the chosen segments at appropriate temporal positions to produce a seamless edit. 
To model this process in a controlled and evaluable manner, we adopt a multiple-choice QA format, which simplifies the workflow while preserving the core reasoning challenge.
Specifically, a pre-cut clip is provided as the \textit{reference video}, and the mode must select and trim one appropriate segment from a list of candidate videos, then decide whether to insert it at the beginning or the end of the reference.
Formally, this task can be represented as
\begin{equation}
(A, t_s, t_e) = M(V_r, [V_1, V_2, \ldots, V_n]),
\end{equation}
where $M$ denotes the LMM, $V_r$ is the reference video, $[V_1, V_2, \ldots, V_n]$ represents the list of candidate videos, and $(A, t_s, t_e)$ corresponds to the selected video index and its start-end timestamps, respectively. 
This formulation allows for a standardized evaluation of a model's ability to reason across multiple videos, infer editing intent, and perform precise temporal localization. 
Notably, this subtask constitutes the \textit{first benchmark of multi-video reasoning} in the context of real-world video editing.

\subsection{Video Collection}
All videos are collected from a public video platform. Due to the different objectives of the two subtasks, we employ different video sourcing strategies for each. 
% \begin{itemize}

\textbf{TechRec} requires videos containing a large number of editing points to generate challenging and diverse questions. 
To build such a corpus, we first leverage GPT-4o to generate thousands of topics covering a wide range of categories (e.g., NBA games, emotional moments in movies, and music performances). 
For each topic, GPT-4o produces multiple corresponding YouTube search queries (e.g., ``\textit{LeBron James highlight moments}'', and ``\textit{touching moments in Interstellar}''), enabling large-scale retrieval of videos across different contexts and genres.

\textbf{OpSim} demands videos with rich editing semantics and naturally interleaved narratives. 
We deliberately select interview videos featuring film actors, as they often alternate between main interviews and complementary B-roll footage (e.g., movie clips), naturally exhibiting cross-references, emotional reinforcement, and narrative alignment. 
We curate samples from high-quality YouTube channels such as \textit{60 Minutes} and \textit{GQ Iconic Characters} to ensure both diversity and stylistic consistency.
% \end{itemize}
Finally, all collected videos are automatically screened using Gemini-2.5-Pro~\cite{google2025gemini25pro}, which filters out samples with poor editing quality or unstable composition, ensuring that the dataset maintains a high level of visual integrity and narrative coherence.

% To ensure the inclusion of high-quality edited content, we first generate specific \textit{video editing topics} tailored to different video genres (e.g., movies, TV shows, interviews, etc.) using GPT-4o. For example, for the movie domain, one topic might be \textit{“The Evolution of James Bond: From Sean Connery to Daniel Craig”}. Based on these topics, we automatically generate thousands of YouTube search queries, such as \textit{“Daniel Craig James Bond best moments”}. For each query, we retain the top five search results to construct the initial video pool. Subsequently, we employ Gemini-2.5-Pro to assess the editing quality of each video, filtering out those that exhibit poor editing craftsmanship. This semi-automated pipeline allows us to efficiently curate a diverse, high-quality collection of real-world edited videos across multiple domains.

\subsection{Annotation Process}
As illustrated in Figure~\ref{fig:annotation}, we design two dedicated annotation pipelines for the two subtasks in \vebenc, each involving a three-round human–AI collaborative annotation and verification process to ensure accuracy and consistency.

\noindent\textbf{Technique Recognition}
For the first subtask, annotators carefully inspect each fine-cut video to locate all transition points (timestamps) and label them with one or more of the seven defined editing techniques: \textit{L-Cut}, \textit{J-Cut}, \textit{Jump Cut}, \textit{Smash Cut}, \textit{Cutaway}, \textit{Invisible Cut}, and \textit{Match Cut}.
Each annotation undergoes three verification rounds: 
1) an initial manual pass for labeling;
2) cross-validation by an independent annotator; and
3) final consistency review using automated checks for audio–visual alignment, particularly for \textit{L-Cut} and \textit{J-Cut}.
This triple-layer validation ensures fine-grained accuracy and high inter-annotator reliability. With this high-quality metadata, we easily generate $5$-option multi-choice QA automatically. The question template can be found in the \textbf{Supp. Sec. 2}.

\noindent\textbf{Operation Simulation}
The second subtask follows a structured multi-stage annotation pipeline that emulates real-world editing workflows, also validated through three iterative annotation–review rounds.
In \textbf{Stage 1}, all videos are first processed by Gemini-2.5-Pro to extract shot-level metadata and scene boundaries.
In \textbf{Stage 2}, given a target B-Roll, annotators identify one or more A-Rolls with strong narrative or semantic connections.
They specify the rationale for each A/B-Roll pair from eight predefined relationship types, and indicate whether the link is grounded in audio, visual, or both modalities. 
% \footnote{\textit{Cause-and-Effect}, \textit{Illustration}, \textit{Contrast/Comparison}, \textit{Emotional/Stylistic Reinforcement}, \textit{Example/Case Study}, \textit{Contextual Background}, \textit{Flashback/Archival Reference}, and \textit{Symbolic/Metaphorical Link}.}
% \dd{how we define 8 relationship types? add the reference. What is the purpose to destingush 8 types?}
In \textbf{Stage 3}, annotators locate the correct source video and its timestamps, then design three semantically plausible distractors (e.g., another scene from the same film, a similar scene from a different movie, or clips sharing the same actor or tone).
Finally, the questions are automatically generated in \textbf{Stage 4} given the template, and in \textbf{Stage 5} annotators perform question–answer validation, namely reviewing all candidate clips, re-selecting the best-fitting option, and providing timestamp-level justifications for any ambiguous cases. 
This three-round annotation pipeline ensures both semantic integrity and reasoning depth, yielding a high-quality dataset that faithfully captures authentic multi-video editing cognition. More details of the annotation can be found in the \textbf{Supp. Sec. 2}.

\subsection{Statistics} %\dd{This section lacks many details. Based on the statistics figure, we need to conclude some insights of our datasets different from others.}
\label{sec:statistics}
\vebenc comprises a total of $3.9K$ edited videos spanning $257$ hours, paired with $3,080$ human-verified QA instances, making it one of the most comprehensive resources for studying multimodal reasoning in real-world video editing.
The dataset is divided into two complementary subtasks (\textbf{TechRec} and \textbf{OpSim}), each capturing distinct aspects of human editing cognition.

\begin{figure}[t]
\centering

\begin{minipage}{0.48\linewidth}
    \centering
    \includegraphics[width=\linewidth]{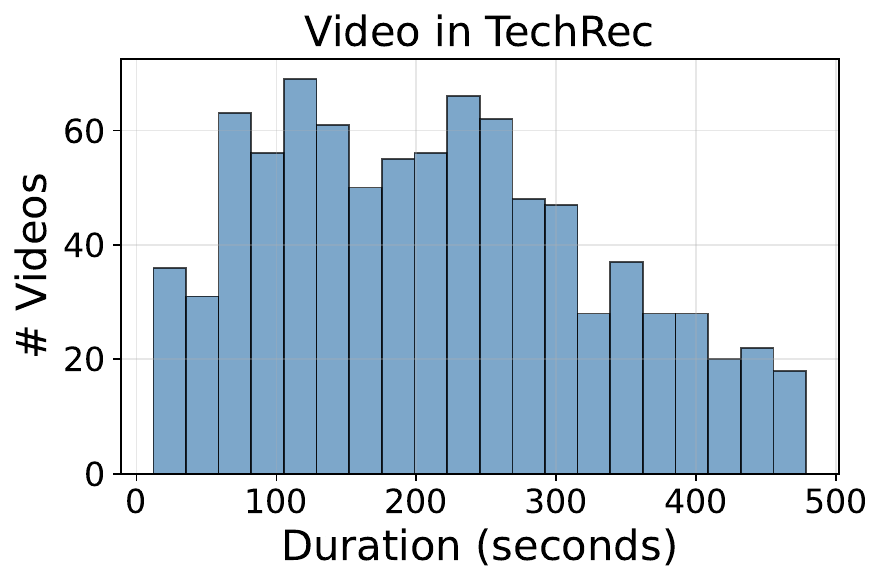}
    % \caption*{(a) Tech video duration distribution.}
\end{minipage}
\hfill
\begin{minipage}{0.48\linewidth}
    \centering
    \includegraphics[width=\linewidth]{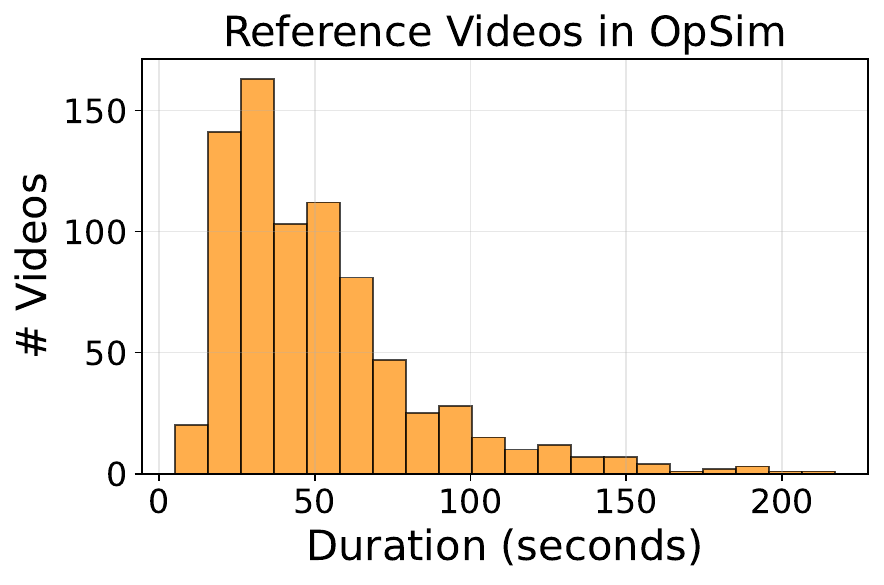}
    % \caption*{(b) Reference video duration distribution.}
\end{minipage}

\vspace{0.4em}

\begin{minipage}{0.48\linewidth}
    \centering
    \includegraphics[width=\linewidth]{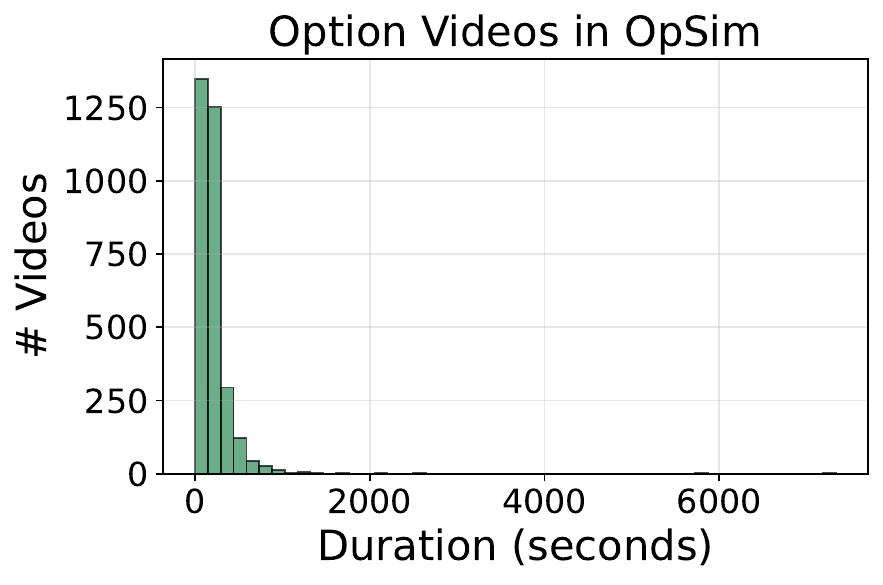}
    % \caption*{(c) Option video duration distribution.}
\end{minipage}
\hfill
\begin{minipage}{0.48\linewidth}
    \centering
    \includegraphics[width=\linewidth]{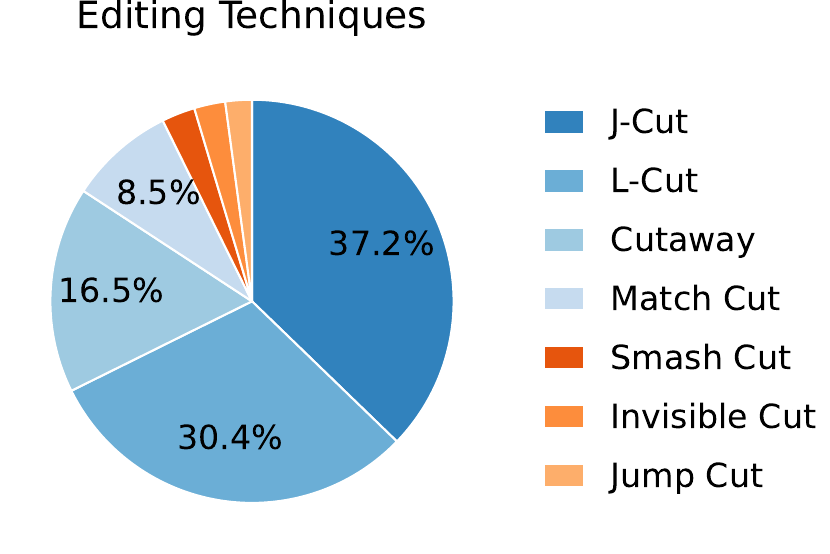}
    % \caption*{(d) Editing Tech Distribution.}
\end{minipage}

\caption{Distributions of video duration and editing techniques in \vebenc. The long-duration videos and rich coverage of diverse editing techniques demonstrate the complexity of \vebenc.}
\label{fig:vebench_stats}
\vspace{-3mm}
\end{figure}

For TechRec, the dataset contains $885$ long-form videos and corresponding QA pairs.
Each video averages $213.6$ seconds in length and includes $3.3$ annotated editing cuts, covering a broad spectrum of styles and transition patterns (Figure~\ref{fig:vebench_stats}).
This subtask emphasizes the perception of editing rhythm, temporal coherence, and multimodal synchronization, posing a significant challenge beyond conventional shot-boundary or action-recognition tasks.

For OpSim, we collect $783$ reference videos and $2,217$ unique candidate videos, forming multi-video QA settings that mimic real-world editing workflows.
The reference videos average $51.6$ seconds, while the candidate videos are substantially longer ($\sim228.5$ seconds), providing rich temporal context for cross-video reasoning and segment-level grounding.
This design requires models to reason across multiple sources, evaluate narrative coherence, and make operational decisions similar to professional editors.

Overall, \vebenc not only achieves large-scale coverage but also embodies multi-level complexity, ranging from fine-grained temporal perception to high-level reasoning over narrative intent.

% \begin{figure}[t]
%     \centering
%     % --- Top figure: video duration ---
%     \includegraphics[width=0.5\textwidth]{figures/video_duration_distribution.pdf}
%     \vspace{4mm}
%     % --- Bottom figure: cut type ---
%     \includegraphics[width=0.3\textwidth]{figures/cut_type_distribution.pdf}
%     \caption{
%         (Top) Histogram of video duration distributions across technique, reference, and option videos. 
%         (Bottom) Cut-type distribution showing the proportion of each editing type in the dataset.
%     }
%     \label{fig:dataset_stats}
% \end{figure}

\section{Experiment}
% 4.1 setting : model, setting, metric
% 4.2 Quantitative Analysis
% 4.3 Quanlitative 
\begin{table*}[t]
\centering
\caption{Evaluation results on \vebenc. ``TechRec'' denotes the \textit{Technique Recognition} subtask, and ``OpSim'' denotes the \textit{Operation Simulation} subtask, where ``FS'' represents \textit{Footage Selection}. The overall score is the average of TechRec and OpSim-FS. The \textbf{bold text} indicates the best performance.}
\resizebox{1\textwidth}{!}{
\begin{tabular}{@{}lllccccccc@{}} % <-- now 11 columns total
\toprule
\multirow{2}{*}{\textbf{Models}} & 
\multirow{2}{*}{\textbf{LLM}} & 
\multirow{2}{*}{\textbf{Size}} &
\multicolumn{2}{c}{\textbf{Overall}} &
\multirow{2}{*}{\textbf{TechRec}} &
\multicolumn{2}{c}{\textbf{OpSim-FS}} &
\multicolumn{2}{c}{\textbf{OpSim}} \\
\cmidrule(lr){4-5} \cmidrule(lr){7-8} \cmidrule(lr){9-10}
 & & &
\textbf{w/ subs} & \textbf{w/o subs} &
 &
\textbf{w/ subs} & \textbf{w/o subs} &
\textbf{w/ subs} & \textbf{w/o subs} \\
\midrule

\rowcolor{gray!10}\multicolumn{10}{c}{\textit{Proprietary Models}} \\
GPT-4o~\cite{gpt4o2024} & -- & -- & 33.29\% & 32.97\% & 24.68\% & 41.89\% & \textbf{41.25\%} & 0.00 & 0.00 \\
Gemini-2.5-Pro~\cite{google2025gemini25pro} & -- & -- & \textbf{39.55\%} & -- & \textbf{34.65\%} & \textbf{44.44\%} & -- & \textbf{0.11} & -- \\

\midrule
\rowcolor{gray!10}\multicolumn{10}{c}{\textit{Open-Source Models}} \\
Qwen3-VL-32B-Instruct~\cite{qwen3vl} & Qwen3~\cite{qwen3} & 32B & 34.36\% & \textbf{34.36\%} & 28.99\% & 39.72\% & 39.72\% & 0.00 & 0.00 \\
Qwen3-VL-30B-A3B-Instruct~\cite{qwen3vl} & Qwen3~\cite{qwen3} & 30B & 28.84\% & 31.65\% & 28.30\% & 29.37\% & 34.99\% & 0.00 & 0.00 \\
Qwen3-VL-8B-Instruct~\cite{qwen3vl} & Qwen3~\cite{qwen3} & 8B & 32.39\% & 33.86\% & 28.64\% & 36.14\% & 39.08\% & 0.00 & 0.00 \\
Qwen3-VL-4B-Instruct~\cite{qwen3vl} & Qwen3~\cite{qwen3} & 4B & 30.20\% & 30.14\% & 28.60\% & 31.80\% & 31.67\% & 0.00 & 0.00 \\
Qwen2.5-VL-7B-Instruct~\cite{Qwen2.5-VL} & Qwen2.5~\cite{qwen25} & 7B & 28.74\% & 27.79\% & 27.47\% & 30.01\% & 28.10\% & 0.00 & 0.01 \\
InternVL3-14B~\cite{zhu2025internvl3} & Qwen2.5~\cite{qwen25} & 14B & 29.12\% & 28.55\% & 27.08\% & 31.16\% & 30.01\% & 0.00 & 0.01 \\
InternVL3-8B~\cite{zhu2025internvl3} & Qwen2.5~\cite{qwen25} & 7B & 27.99\% & 26.08\% & 25.21\% & 30.78\% & 26.95\% & 0.00 & 0.00 \\
% VITA-1.5 \\
% VideoLLaMA3 \\
% Qwen3-Omni-30B-A3B \\

\bottomrule
\end{tabular}
}
\label{tab:main}
\vspace{-3pt}
\end{table*}

\subsection{Evaluation Settings}
We conducted extensive experiments to analyze the performance of current LMMs and demonstrate the value of the new benchmark.
Specifically, we evaluate two state-of-the-art proprietary models—Gemini-2.5-Pro~\cite{google2025gemini25pro} and GPT-4o~\cite{gpt4o2024}, which currently achieve strong results on existing video understanding benchmarks.
In addition, we comprehensively assess $7$ open-source LMMs from the Qwen-VL~\cite{Qwen2.5-VL,qwen3vl} and InternVL families~\cite{zhu2025internvl3}.

\noindent\textbf{Metrics.}
For \textbf{TechRec}, we use \textit{accuracy} as the evaluation metric following the standard protocol for multi-choice VideoQA tasks~\cite{fu2024videomme,wu2024longvideobench,deng2025scivideobench}.
For \textbf{OpSim}, we report two metrics:
1) OpSim-FS (Footage Selection Accuracy), the accuracy of selecting the correct video option, and
2) Full-task performance, which evaluates both selection and temporal localization.
For the latter, we compute a conditioned temporal Intersection-over-Union (tIoU\textsubscript{cond}), \ie,
\begin{equation}
\text{tIoU}_{\text{cond}} = 
\begin{cases}
\text{tIoU}(p, g), & \text{if } A_p = A_g, \\
0, & \text{otherwise},
\end{cases}
\end{equation}
where $\text{tIoU}(p, g)$ denotes the temporal IoU between the predicted and ground-truth segments, and $A_p$, $A_g$ are the predicted and ground-truth video choices, respectively.
This conditioned formulation penalizes models that select the wrong video source, thereby aligning evaluation with realistic editing workflows that require both correct footage selection and temporal precision.

\noindent\textbf{Implementation Details.}
For \textbf{TechRec}, we follow the standard VideoQA evaluation setup~\cite{fu2024videomme,wu2024longvideobench,deng2025scivideobench}. 
For \textbf{OpSim}, which involves multi-video reasoning, we note that most open-source LMMs (except Gemini-2.5-Pro~\cite{google2025gemini25pro}) do not natively support simultaneous multi-video input.
To address this limitation, we concatenate all candidate videos into a single continuous sequence and insert on-screen text prompts to delineate the reference clip from each option (see Figure~\ref{fig:v_prompt}(a)).
This unified prompting allows fair comparison across models while preserving temporal coherence.
Frame sampling for each model follows its official default configuration.
Furthermore, to analyze the contribution of auditory cues, we test two variants for each model (with and without subtitles) on the OpSim task, enabling a controlled comparison of multimodal grounding effects.

\subsection{Quantitative Results}
As summarized in Table~\ref{tab:main}, current LMMs exhibit severe performance gaps on both subtasks, revealing the fundamental difficulty of real-world video editing reasoning.
While Gemini-2.5-Pro~\cite{google2025gemini25pro} consistently outperforms all other models, its advantage mainly stems from its stronger audio–visual perception and temporal reasoning capabilities rather than true editing cognition.
Even so, its overall performance remains far below the level required for practical editing applications.

For \textbf{TechRec}, audio cues prove essential, especially for detecting J-/L-Cuts and Smash Cuts that depend on cross-scene sound continuity.
Models without audio understanding have poor performance, indicating a lack of multimodal synchronization awareness.
Gemini-2.5-Pro~\cite{google2025gemini25pro} achieves the highest score, yet still struggles with fine-grained boundary detection and complex transitions in long videos.

For \textbf{OpSim}, the challenge becomes even more pronounced.
When restricted to footage selection (OpSim-FS), Gemini-2.5-Pro~\cite{google2025gemini25pro} performs comparably to GPT-4o~\cite{gpt4o2024} and Qwen3-VL-32B-Instruct~\cite{qwen3vl}, suggesting partial reasoning ability across multiple candidate clips.
However, once the task extends to temporal localization, nearly all models fail completely, achieving tIoU scores close to zero.
Only Gemini-2.5-Pro~\cite{google2025gemini25pro} produces non-trivial results, demonstrating limited yet meaningful temporal grounding.

Overall, these findings confirm that existing LMMs are still far from human-level editing reasoning.
The large performance gap highlights the unique challenge posed by \vebenc, where success requires joint reasoning over audio–visual cues, temporal continuity, and cross-video narrative coherence—capabilities that remain beyond the reach of current models.

\noindent\textbf{The Impact of Subtitles.}
We further investigate the effect of subtitles (\textit{w/ subs}) on the operation simulation task.
As shown in Table~\ref{tab:main}, the presence of subtitles consistently improves model performance across both subtasks.
For example, in the OpSim-FS setting, Gemini-2.5-Pro~\cite{google2025gemini25pro} improves from $41.25\%$ to $44.44\%$, and Qwen3-VL-8B-Instruct~\cite{qwen3vl} from $38.52\%$ to $39.08\%$.
This moderate yet consistent gain indicates that subtitles provide additional semantic grounding by explicitly linking dialogue or narration to visual context, reinforcing temporal alignment across shots.
In essence, subtitles act as a lightweight cross-modal bridge, compensating for LMMs' limited ability to infer narrative flow directly from raw audio–visual cues.
\begin{figure}[t]
  \centering
   \includegraphics[width=1\linewidth]{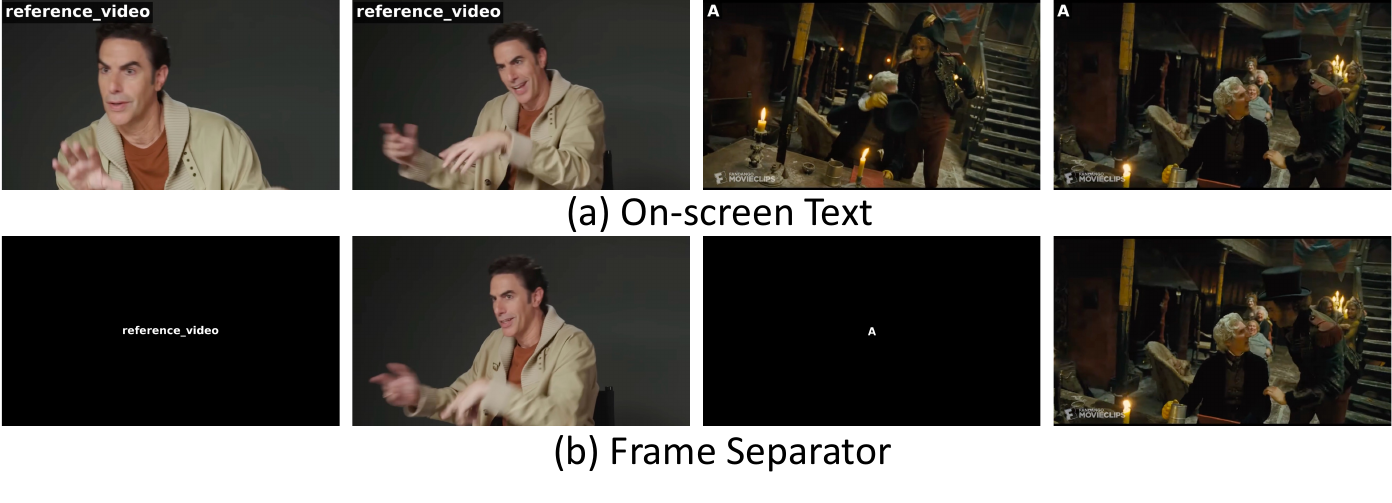}
   \caption{Examples of two different visual prompts in the stitched videos: (a) On-screen Text and (b) Frame Separator.}
   \label{fig:v_prompt}
   \vspace{-5pt}
\end{figure}

\begin{figure*}[t]
  \centering
   \includegraphics[width=.95\linewidth]{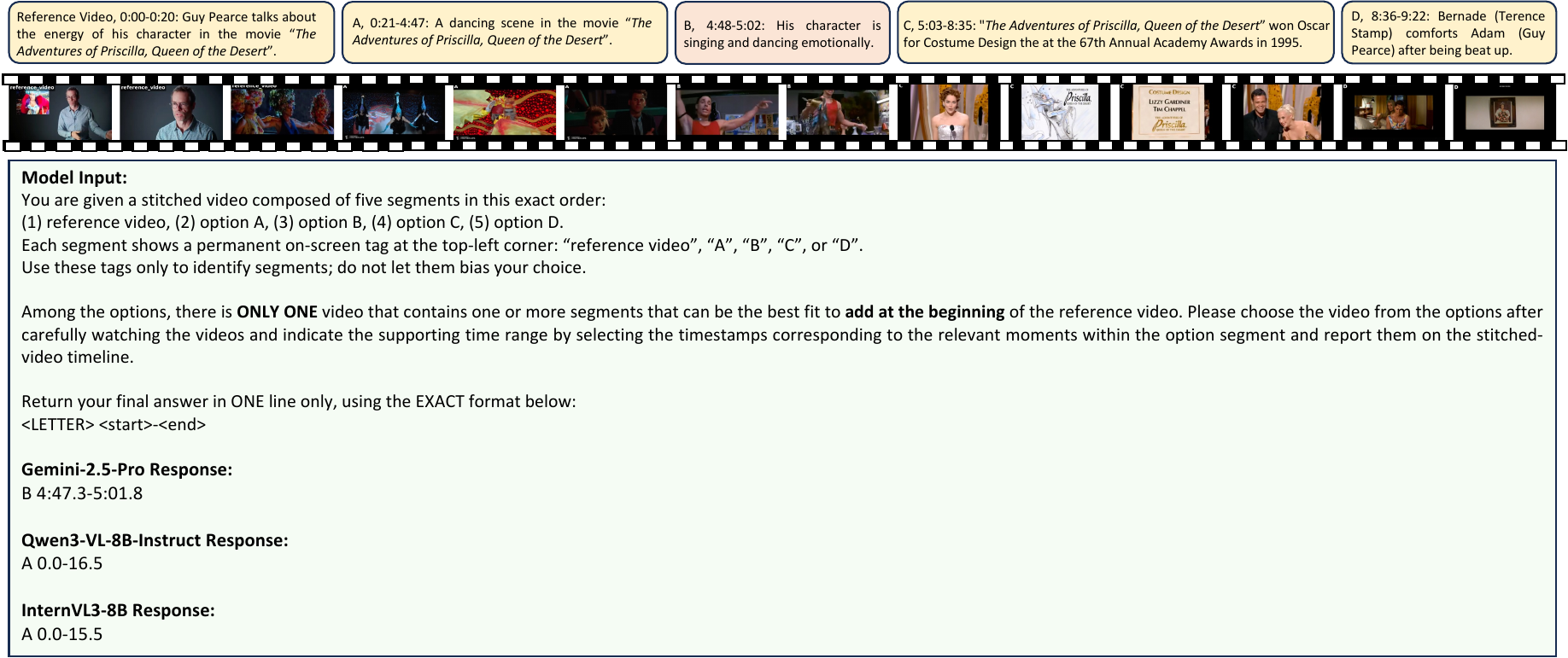}
   \caption{Qualitative example of Gemini-2.5-Pro~\cite{google2025gemini25pro}, Qwen3-VL-8B-Instruct~\cite{qwen3vl}, and InternVL3-8B~\cite{zhu2025internvl3}. }
   \label{fig:visualization}
   \vspace{-3pt}
\end{figure*}

\noindent\textbf{The Impact of Visual Prompts.}
We also analyze the influence of \textit{visual prompting strategies} on model reasoning in the OpSim task.
In Figure \ref{fig:v_prompt}, beyond the default on-screen text (OT) used to mark reference and candidate clips, we introduce a frame separator (SP), a blank transition frame inserted between video segments, to isolate visual boundaries.
Based on Table~\ref{tab:v_prompt}, both proprietary and open-source models exhibit clear performance degradation when using separators instead of on-screen text.
For instance, the accuracy of Qwen3-VL-8B-Instruct~\cite{qwen3vl} drops by over $9\%$, and Gemini-2.5-Pro~\cite{google2025gemini25pro} declines from $44.44\%$ to $40.23\%$.
This demonstrates that continuous visual continuity cues (as provided by on-screen text) are crucial for maintaining temporal coherence and contextual reasoning, whereas abrupt separators disrupt the model's understanding of narrative flow.

Overall, these analyses reveal that current LMMs remain heavily reliant on explicit linguistic or visual markers, such as subtitles and on-screen text, to organize temporal and semantic context. It indicates their lack of robust internal representation for video continuity and editing logic.
\begin{table}[t]
\centering
\caption{The comparison of different visual prompts used in the operation simulation task. ``OT'' indicates on-screen text, and ``SP'' indicates the frame separator.}
\resizebox{0.45\textwidth}{!}{
\begin{tabular}{@{}lcccc@{}}
\toprule
\textbf{Models} &
\multicolumn{2}{c}{\textbf{OpSim-FS}} &
\multicolumn{2}{c}{\textbf{OpSim-SL}} \\
\cmidrule(lr){2-3} \cmidrule(lr){4-5}
& \textbf{w/ OT} & \textbf{w/ SP} & \textbf{w/ ST} & \textbf{w/ SP} \\
\midrule
Gemini-2.5-Pro~\cite{google2025gemini25pro} & 44.44\% & 40.23\% & 0.11 & 0.07 \\
Qwen3-VL-8B-Instruct~\cite{qwen3vl} & 39.08\% & 30.01\% & 0.00 & 0.00 \\
\bottomrule
\end{tabular}
}
\label{tab:v_prompt}
\vspace{-5pt}
\end{table}

% \noindent\textbf{Timestamp Prediction Bias.}
% The timestamp predictions of different models exhibit distinct biases, which help explain the large performance disparity in the OpSim-SL task. As shown in Figure~\ref{fig:pred-timestamp-all}, both Qwen3-VL-8B-Instruct~\cite{qwen3vl} and InternVL3-8B~\cite{zhu2025internvl3} demonstrate severe biases toward predicting timestamps near the beginning of the video (the average duration of the stitched video is $941.67s$), indicating a failure to perform fine-grained temporal reasoning. This behavior leads to almost all predictions being misaligned with the ground truth segments, resulting in tIoU values close to 0 as shown in Table~\ref{tab:main}. In contrast, Gemini-2.5-Pro~\cite{google2025gemini25pro} produces more diverse and distributed timestamp predictions across the entire video, reflecting a better understanding of temporal continuity and localized event boundaries. Consequently, Gemini-2.5-Pro~\cite{google2025gemini25pro} achieves a non-trivial tIoU of $0.11$, showing a relatively stronger ability to solve real video editing problem.
\vspace{-4pt}
\subsection{Visualization}
We present an example of the \textbf{OpSim} task in Figure~\ref{fig:visualization}. The reference video is an interview segment in which the actor discusses the energy of his character in the movie \textit{The Adventures of Priscilla, Queen of the Desert}, followed by a B-roll dancing scene. The goal of the task is to identify the most suitable segment to insert \textit{before} the reference video. Intuitively, a coherent editing sequence would begin with a scene that visually introduces the character’s energy and emotion, making option \textbf{B} the best choice. While options \textbf{A}, \textbf{C}, and \textbf{D} are thematically related, none establish a logical narrative flow when placed before the interview. In this case, Gemini-2.5-Pro~\cite{google2025gemini25pro} produces a satisfactory prediction by correctly selecting option B and localizing the precise timestamp, whereas Qwen3-VL-8B-Instruct~\cite{qwen3vl} and InternVL3-8B~\cite{zhu2025internvl3} generate irrelevant predictions, indicating limited temporal and contextual reasoning ability. More examples can be found in the \textbf{Supp. Sec. 6}.

% \dd{need to add another discussion section to provide insights and potential research directions.}

\subsection{Limitations of Existing LMMs on Video Editing}
Our evaluation reveals three major factors limiting current LMMs in real-world video editing reasoning:
\begin{figure}[t]
    \centering
    \includegraphics[width=1\linewidth]{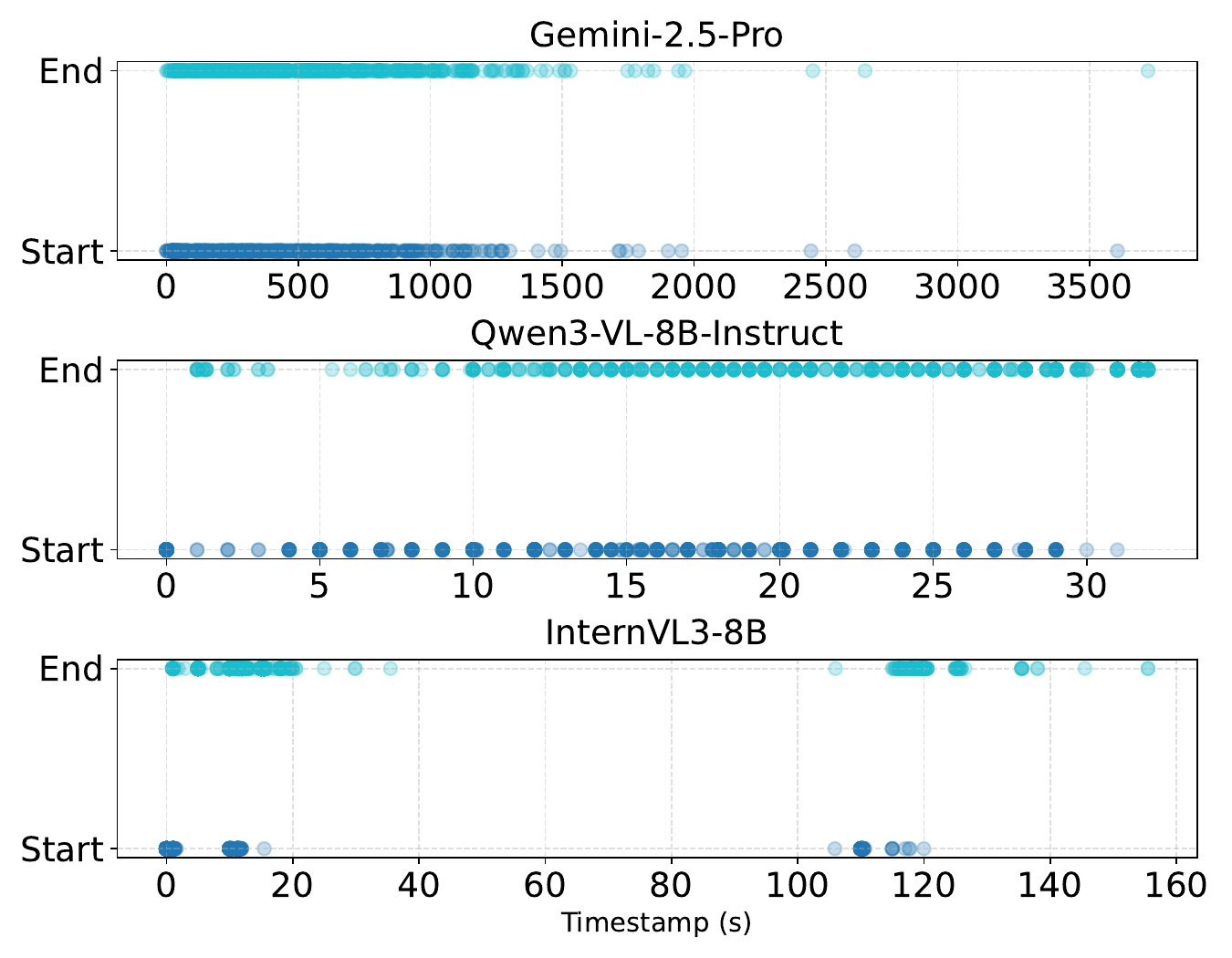}
    \caption{Predicted timestamp distributions in OpSim for Gemini-2.5-Pro~\cite{google2025gemini25pro}, Qwen3-VL-8B-Instruct~\cite{qwen3vl}, and InternVL3-8B~\cite{zhu2025internvl3}. It is obvious that Open-source models tend to predict early timestamps in this task, while Gemini-2.5-Pro showcases a reasonable distribution. The average duration of the video is 941.67s.}
    \label{fig:pred-timestamp-all}
\end{figure}

\begin{itemize}
    \item \textbf{Weak audio–visual integration.} 
    Most models treat vision and audio independently, failing to capture synchronized cues essential for techniques such as J-/L-Cuts.
    Without cross-modal alignment, they cannot reason about temporal or emotional continuity across scenes.

    \item \textbf{Limited temporal capacity.} 
    Many LMMs handle only short sequences (e.g., InternVL3~\cite{zhu2025internvl3} supports $64$ frames), missing fine-grained motion, transitions, and stylistic changes that real editing requires across hundreds of frames.
    In contrast, real-world editing demands fine-grained temporal sensitivity, requiring models to reason over hundreds of frames and seamlessly track semantic evolution across scenes.

    \item \textbf{Inaccurate temporal localization.} 
    As shown in Figure~\ref{fig:pred-timestamp-all}, in the Operation Simulation task, open-source models always predict timestamps near video beginnings (average duration $941.67$s), leading to near-zero tIoU scores in Table~\ref{tab:main}.
    Even Gemini-2.5-Pro~\cite{google2025gemini25pro}, with a non-trivial tIoU of $0.11$, remains far from the precision needed for practical workflows.
\end{itemize}
These limitations highlight the diagnostic strength of \vebenc and the urgent need for next-generation models capable of unifying multimodal understanding with goal-driven operational reasoning.
% 1. audio processing ability 
% 2. model capacity (InternVL 64 frames)
% 3. temporal loc, time prediction bias 

\section{Conclusion}
We presented \textbf{\vebenc}, a comprehensive benchmark for video editing that unifies both editing \textit{knowledge understanding} and \textit{operational reasoning}.
Unlike prior benchmarks focusing on perception within short single-clip contexts, \vebenc highlights the dual nature of editing that requires not only recognition of editing techniques but also reasoning over multi-video composition and real-world editing workflows.
The benchmark comprises two complementary subtasks: Video Editing Technique Recognition, which evaluates the perception and comprehension of editing styles, and Video Editing Operation Simulation, which challenges models to select and temporally localize suitable clips across multiple videos.
Extensive evaluations on both proprietary and open-source LMMs reveal a significant gap between current model capabilities and the demands of authentic editing reasoning, even for state-of-the-art systems such as Gemini-2.5-Pro.
\vebenc lays the groundwork for next-generation multimodal systems that move beyond passive video understanding toward human-level editing intelligence and creative reasoning.

{
    \small
    \bibliographystyle{ieeenat_fullname}
    \bibliography{main}
}

\appendix

\section{Video Editing Background Knowledge}
\label{supp:tech}

\subsection{Editing Technique Definition}

In this section, we define and illustrate the seven fundamental video editing techniques~\cite{frierson2018film} included in \textbf{\vebenc}. Each technique is characterized by its modality (visual or audiovisual) and its unique function in shaping cinematic rhythm, emotion, and continuity.

% =============================
% J-CUT
% =============================
\noindent\textbf{J-Cut (Audio Leads Video)} \quad [Modality: A+V] \\
The audio from the next scene begins before the visual cut occurs, producing a smooth and anticipatory transition that connects scenes through sound continuity. Named after the shape of the letter “J,” this technique allows audio to lead the visual transition.

\noindent{Movie Example:} \textit{The Wolf of Wall Street} (2013), the well-known “Mmm-hmm” conversation between McConaughey and DiCaprio.  
\href{https://www.youtube.com/watch?v=eyH-a964kAs&t=10}{https://www.youtube.com/watch?v=eyH-a964kAs (0:10)}
% \begin{figure}[h]
%     \centering
%     \includegraphics[width=0.7\linewidth]{figures/examples/jcut_example1.pdf}\\[3pt]
%     \includegraphics[width=0.7\linewidth]{figures/examples/jcut_example2.pdf}
%     \caption*{Visual examples of \textbf{J-Cut}.}
% \end{figure}

% =============================
% L-CUT
% =============================
\noindent\textbf{L-Cut (Audio Follows Video)} \quad [Modality: A+V] \\
The audio from the current scene continues to play after the video cuts to the next one, maintaining auditory continuity and emotional flow. It mirrors the letter “L,” where the audio extends beyond the visual boundary.

\noindent{Movie Example:} \textit{The Tree of Life} (2011) — 
Young Jack opens and closes a door, and the creaking sound continues even after the visual cut, creating a lingering auditory bridge.\\
\href{https://www.youtube.com/watch?v=eyH-a964kAs&t=53}{https://www.youtube.com/watch?v=eyH-a964kAs (0:53)}

% \begin{figure}[h]
%     \centering
%     \includegraphics[width=0.7\linewidth]{figures/examples/lcut_example1.pdf}\\[3pt]
%     \includegraphics[width=0.7\linewidth]{figures/examples/lcut_example2.pdf}
%     \caption*{Visual examples of \textbf{L-Cut}.}
% \end{figure}

% =============================
% SMASH CUT
% =============================
\noindent\textbf{Smash Cut (Breaking Continuity)} \quad [Modality: V only] \\
A sudden, jarring cut between two dramatically different scenes in tone, sound, or content. Smash cuts heighten contrast and shock, emphasizing emotional or narrative breaks.

\noindent{Movie Example:} \textit{Reservoir Dogs} (1992) — 
A sudden alarm blast cuts sharply into the scene as Mr.\ Pink sprints down the street clutching the stolen diamonds while gunfire erupts behind him, creating a jarring, high-intensity transition.\\
\href{https://www.youtube.com/watch?v=Q5Y5iqVNrls&t=43}{https://www.youtube.com/watch?v=Q5Y5iqVNrls (0:43–0:49)}

% \begin{figure}[h]
%     \centering
%     \includegraphics[width=0.7\linewidth]{figures/examples/smashcut_example1.pdf}\\[3pt]
%     \includegraphics[width=0.7\linewidth]{figures/examples/smashcut_example2.pdf}
%     \caption*{Visual examples of \textbf{Smash Cut}.}
% \end{figure}

% =============================
% JUMP CUT
% =============================
\noindent\textbf{Jump Cut (Time Condensation)} \quad [Modality: V only] \\
A discontinuous edit that skips forward in time within the same scene while maintaining spatial coherence. Jump cuts condense time, speed up pacing, or create rhythmic tension in visual storytelling.

\noindent{Movie Example:} \textit{Breathless} (1960), directed by Jean-Luc Godard — 
The protagonist walks along the street and abruptly “jumps forward” in the same motion path, showcasing one of the earliest iconic uses of the jump cut.\\
\href{https://www.youtube.com/watch?v=_fcRj0SXYh8&t=59}{https://www.youtube.com/watch?v=\_fcRj0SXYh8 (0:59–1:13)}

% \begin{figure}[h]
%     \centering
%     \includegraphics[width=0.7\linewidth]{figures/examples/jumpcut_example1.pdf}\\[3pt]
%     \includegraphics[width=0.7\linewidth]{figures/examples/jumpcut_example2.pdf}
%     \caption*{Visual examples of \textbf{Jump Cut}.}
% \end{figure}

% =============================
% CUTAWAY
% =============================
\noindent\textbf{Cutaway (Narrative Supplementation)} \quad [\textit{Modality: V only}]\\
A cutaway momentarily departs from the main action to show a related shot—such as a reaction, an important detail, or an environmental element—before returning to the primary scene. Cutaways enrich narrative context, emphasize key information, build tension, or provide a smooth visual bridge between edits.

\noindent{Movie Example:} \textit{An American Werewolf in London} (1981) — 
During the transformation sequence, the film abruptly cuts from the violent werewolf attack to a cheerful cartoon of Mickey Mouse, then cuts back to the attack.\\
\href{https://www.youtube.com/watch?v=WrIwfImLXOA}{https://www.youtube.com/watch?v=WrIwfImLXOA}

% \begin{figure}[h]
%     \centering
%     \includegraphics[width=0.7\linewidth]{figures/examples/cutaway_example1.pdf}\\[3pt]
%     \includegraphics[width=0.7\linewidth]{figures/examples/cutaway_example2.pdf}
%     \caption*{Visual examples of \textbf{Cutaway}.}
% \end{figure}

% =============================
% MATCH CUT
% =============================
\noindent\textbf{Match Cut (Visual Matching)} \quad [Modality: V only] \\
A transition that connects two visually or thematically similar shots to create a symbolic or continuous relationship. Match cuts enhance visual coherence and conceptual association between scenes.

\noindent{Movie Example:} \textit{2001: A Space Odyssey} (1968) — 
A bone tossed into the air cuts seamlessly to a spacecraft occupying the same position in the frame, creating one of the most iconic match cuts in cinema history.\\
\href{https://www.youtube.com/watch?v=iGuv1G3-f7w&t=27}{https://www.youtube.com/watch?v=iGuv1G3-f7w (0:27–0:31)}

% \begin{figure}[h]
%     \centering
%     \includegraphics[width=0.7\linewidth]{figures/examples/matchcut_example1.pdf}\\[3pt]
%     \includegraphics[width=0.7\linewidth]{figures/examples/matchcut_example2.pdf}
%     \caption*{Visual examples of \textbf{Match Cut}.}
% \end{figure}

% =============================
% INVISIBLE CUT
% =============================
\noindent\textbf{Invisible Cut (Creating Artificial Continuity)} \quad [\textit{Modality: V only}]\\
An invisible cut is a seamless edit that hides the transition between two shots, creating the illusion of a continuous, unbroken take. The cut is typically concealed through fast camera motion, object occlusion, or brief darkness (such as passing behind a wall or a whip pan), preserving immersion and narrative flow.

\noindent{Movie Example:} \textit{1917} (2019) — 
The film extensively employs invisible cuts to maintain the appearance of a single continuous shot.\\
\href{https://www.youtube.com/watch?v=ZAQoY3ioci0}{https://www.youtube.com/watch?v=ZAQoY3ioci0}

\subsection{Video Editing Terminology in Interview Video}

Interview-style videos commonly rely on two foundational components of professional editing: \textbf{A-Roll} and \textbf{B-Roll}. Understanding these elements and how they interact are essential for analyzing and reconstructing real-world editing workflows.

\noindent\textbf{A-Roll} refers to the primary interview footage, typically featuring the subject speaking directly to the camera or to the interviewer. It carries the core narrative, emotional delivery, and informational content. A-Roll establishes context, provides continuity, and anchors the structure of the video.

\noindent\textbf{B-Roll} consists of supplemental visual material intercut with the A-Roll. In interview videos, B-Roll may include movie scenes, archival footage, behind-the-scenes clips, environment shots, reaction shots, or thematic imagery. B-Roll serves multiple editorial purposes: enriching narrative context, illustrating references made by the interviewee, smoothing transitions, reinforcing emotional tone, or providing pacing and visual variety. 

Modern cinematic interview formats (e.g., 60 Minutes, Variety, GQ) rely heavily on purposeful B-Roll selection. Editors choose and position these clips with careful intent, to visualize abstract ideas, highlight key roles or achievements, foreshadow upcoming discussion, or maintain viewer engagement through dynamic visual rhythm. In \vebenc, these elements are essential for constructing the Operation Simulation task. To solve OpSim effectively, a model must understand how A-Roll and B-Roll segments are connected, why specific footage is selected, and how each clip contributes to the broader storytelling logic. This requires reasoning not only about visual-semantic alignment but also about narrative intent, temporal continuity, and the editorial motivations behind placing a particular B-Roll segment at a precise moment in the interview.

\section{Annotation Details}
\label{supp:annotation}

\subsection{TechRec QA Template}
\label{sec:techrec_template}

For the \textbf{Technique Recognition (TechRec)} task, each question is generated based on the temporal and structural characteristics of the editing technique in the annotated video segment. We design three categories of question templates to reflect different annotation conditions, ensuring coverage of unique, repeated, and long-duration editing events. Below, we summarize the three rule types and provide multiple natural-language templates for each.

\vspace{4pt}
\noindent\textbf{Unique Occurrence (Single Cut).}  
If the target editing technique appears exactly once in the video, the question focuses on temporal localization of that single event.  
Example templates:
\begin{itemize}
    \item \textit{``When does the \{\texttt{CUT\_TYPE}\} occur in the video?''}
    \item \textit{``At what time does the \{\texttt{CUT\_TYPE}\} happen?''}
    \item \textit{``Locate the moment when the \{\texttt{CUT\_TYPE}\} appears in the video.''}
    \item \textit{``During which timestamp does the video apply a \{\texttt{CUT\_TYPE}\}?''}
\end{itemize}

\vspace{6pt}
\noindent\textbf{requent Cuts (Count-Based).}  
If the technique appears repeatedly (four or more times), we ask the model to count the number of occurrences instead of localizing each one.  
Example templates:
\begin{itemize}
    \item \textit{``How many \{\texttt{CUT\_TYPE}\} occur in this video?''}
    \item \textit{``Count how many times the video uses a \{\texttt{CUT\_TYPE}\}.''}
    \item \textit{``What is the total number of \{\texttt{CUT\_TYPE}\} in the video?''}
    \item \textit{``How frequently does the video apply the \{\texttt{CUT\_TYPE}\}?''}
\end{itemize}

\vspace{6pt}
\noindent\textbf{Long-Duration Cut (Transition Range).}  
If the editing technique spans a long temporal range (duration $>$5s), the question asks for identifying the editing type used over a specific interval.  
Example templates:
\begin{itemize}
    \item \textit{``What is the video editing technique used from \{\texttt{START}\} to \{\texttt{END}\}?''}
    \item \textit{``Which editing technique is applied during the segment \{\texttt{SPAN}\}?''}
    \item \textit{``Identify the editing technique used throughout the interval \{\texttt{SPAN}\}.''}
    \item \textit{``What cut type is present in the time range \{\texttt{SPAN}\}?''}
\end{itemize}

\vspace{6pt}
\noindent\textbf{Additional Positional Question Templates.}  
When referring to specific ordinal or positional cuts within the video, we additionally adopt complementary question forms:
\begin{itemize}
    \item \textit{``What is the video editing technique used at the second cut in the video?''}
    \item \textit{``What editing technique is applied at the beginning of the video?''}
    \item \textit{``Which editing technique is used at the final cut of the video?''}
\end{itemize}

These templates collectively ensure diversity in phrasing while maintaining structural clarity, enabling robust evaluation of an LMM's ability to recognize and localize video editing techniques across multiple temporal conditions.

\subsection{OpSim QA Template}
\label{sec:opsim_template}

For the \textbf{Operation Simulation (OpSim)} task, each question is constructed around the temporal relationship between a reference A-Roll segment and its corresponding B-Roll segment that should be inserted into the editing timeline. The reference A-Roll provides contextual information from the interview video, while the target B-Roll represents the correct movie clip that needs to be selected and temporally localized.

To ensure temporal consistency with real editing workflows, we design two complementary question templates depending on whether the target B-Roll should be inserted \textit{after} or \textit{before} the reference segment. The model must not only choose the correct footage from a candidate list, but also determine the precise timestamps that define the best-fitting segment.

\vspace{3pt}
\noindent\textbf{(1) Target B-Roll Appears \textit{After} the Reference Video.}  
When the correct B-Roll follows the reference A-Roll chronologically, the question instructs the model to identify the clip that best continues the current sequence:
\begin{quote}
\textit{``What is the best fit to add at the end of the current video clip? Please choose the footage and determine the corresponding timestamps.''}
\end{quote}

\vspace{3pt}
\noindent\textbf{(2) Target B-Roll Appears \textit{Before} the Reference Video.}  
When the appropriate B-Roll precedes the A-Roll in the original edit, the question instead requests the footage that provides the most coherent lead-in:
\begin{quote}
\textit{``What is the best fit to add at the beginning of the current video clip? Please choose the footage and determine the corresponding timestamps.''}
\end{quote}

\vspace{3pt}
These two templates allow OpSim to capture realistic editing operations such as narrative setup, context reinforcement, and visual continuation. They also ensure that the temporal intent of each editing action is explicitly communicated to the model, enabling accurate evaluation of both footage selection and temporal localization.

\subsection{OpSim Metadata Generation Prompt}
The metadata used for the Operation Simulation task consists of a rich, structured JSON representation that captures both global editing attributes and fine-grained temporal scene annotations of an interview video. The prompt for Gemini-2.5-Pro~\cite{google2025gemini25pro} is shown in Figure~\ref{fig:prompt_p1} and Figure~\ref{fig:prompt_p2}.
As illustrated in the example shown in Figure~\ref{fig:metadata_example}, each video entry contains high-level fields such as \texttt{editing\_quality}, \texttt{interviewee}, and detailed justifications describing the editing style, pacing, and cinematic integration of B-Roll footage. The core of the metadata is the \texttt{timestamp\_breakdown}, a sequential list of short, non-overlapping segments, each annotated with precise start and end times, a scene description, the inferred source movie (if applicable), and a scene type distinguishing A-Roll interviews, B-Roll cinematic inserts, intro/outro graphics, or narration overlays. For B-Roll segments, we additionally record their narrative purpose, a five-level uniqueness rating, a textual justification explaining why the clip is (ir)replaceable, and a set of five search queries that can be used to retrieve the original source footage. Together, this metadata provides a comprehensive temporal decomposition of the video and serves as the foundation for constructing OpSim questions, selecting reference A-Roll, retrieving candidate B-Rolls, and forming meaningful distractors during dataset generation.

\subsection{A/B Roll Pairing Rules}

A key component of constructing the Operation Simulation task is the pairing of A-Roll (interview footage) with B-Roll (movie clips, archival footage, or other illustrative visuals). To ensure that the B-Roll associated with each A-Roll segment is semantically meaningful and reflects real editorial intent, we develop a structured annotation protocol that categorizes the narrative function of each A/B Roll pairing into \textbf{eight editorial relationship types}. These categories capture how professional editors use B-Roll to support, enrich, or shape the interview narrative.

The eight relationship types are defined as follows:

\begin{itemize}
    \item \textbf{Cause-and-Effect} \\
    The B-Roll visualizes a direct outcome or consequence of what is being discussed in the A-Roll. It reinforces logical progression or explains how one event leads to another.

    \item \textbf{Illustration} \\
    The B-Roll provides a literal visual depiction of a concept, object, location, or event referenced verbally in the A-Roll. This is one of the most common uses in interview editing.

    \item \textbf{Contrast / Comparison} \\
    The B-Roll offers a contrasting example or a comparative visual that frames the current discussion—e.g., showing a past role versus a current role, or contrasting moods and styles.

    \item \textbf{Emotional / Stylistic Reinforcement} \\
    The B-Roll amplifies the emotional tone, aesthetic, or atmosphere of the interview segment. This type is often used to build mood or emphasize personality traits.

    \item \textbf{Example / Case Study} \\
    The B-Roll presents a concrete example (e.g., a particular film scene or public event) that supports a general claim or theme mentioned in the A-Roll.

    \item \textbf{Contextual Background} \\
    The B-Roll provides essential background information—historical, geographical, or biographical—that situates the A-Roll content within a broader context.

    \item \textbf{Flashback / Archival Reference} \\
    The B-Roll uses past footage, archival material, or retrospective scenes to recall previous events or earlier phases in the interviewee’s career.

    \item \textbf{Symbolic / Metaphorical Link} \\
    The B-Roll serves a symbolic purpose rather than a literal one. It visually expresses a theme, metaphor, or abstract concept mentioned in the A-Roll.
\end{itemize}

These relationship types are used by annotators when pairing A-Roll and B-Roll segments during dataset creation. By explicitly modeling the editorial intent behind each pairing, \vebenc captures the narrative structure found in professionally edited interviews. This allows the Operation Simulation task to more faithfully reflect real-world editing workflows, in which selecting the ``correct'' footage requires understanding not only visual content but also the underlying storytelling logic.

\onecolumn
\begin{center}
\begin{minipage}{0.96\textwidth}
\begin{tcolorbox}[title=\textbf{Prompt (P1) Example for Video Metadata Generation},
  colback=gray!3, colframe=black!40, boxrule=0.4pt, arc=2pt,
  left=2pt, right=2pt, top=2pt, bottom=2pt,
  width=\textwidth, breakable]

\begin{lstlisting}[style=promptstyle]
You are an expert in film editing and visual media analysis. 
I will give you an interview video, and your job is to perform a professional multi-step analysis.
The final output should be in strict JSON format.

---
### Your Tasks

1. **Editing Quality Assessment** 
- Evaluate the overall editing quality of the interview video. 
- Rate it as one of: "high", "medium", or "low". 
- Consider transitions, pacing, timing to music/speech, and use of overlays.
- Provide a detailed justification in "editing_justification".

2. **Dataset Suitability Decision** 
- Decide whether this video is suitable for inclusion in a multimodal dataset 
  focused on edited cinematic clips. 
- Must include meaningful edits, cinematic framing (16:9), and exclude shorts or raw footage.
- Output "Yes"/"No" in "suitable_for_dataset" and explain in "suitability_justification".

3. **Interviewee Identification** 
- Identify or infer the interviewee's name from context or transcript.

4. **Timestamp-Based Scene Breakdown**
For each scene, return:
- "timestamp_start": e.g., "0:24"
- "timestamp_end": e.g., "0:31"
- "scene_description": visual content summary
- "source_movie": name or None
- "type": "A-Roll", "B-Roll", "Intro Graphic", etc.
- "b_roll_purpose": narrative function (if B-Roll)
- "broll_uniqueness": "Highly unique" or "Highly replaceable" (if B-Roll)
- "broll_uniqueness_justification": reasoning (if B-Roll)
- "source_search_queries": five YouTube/Google queries (if B-Roll)

\end{lstlisting}

\end{tcolorbox}
\vspace{-4pt}
\captionof{figure}{The prompt (P1) used for video analysis in \vebenc metadata generation. 
It asks the model to perform editing-quality evaluation, dataset suitability judgment, interviewee identification,
and timestamp-level scene annotation in strict JSON format.}
\label{fig:prompt_p1}
\end{minipage}
\end{center}
\twocolumn
% \clearpage

\clearpage
\onecolumn
\begin{center}
\begin{minipage}{0.96\textwidth}
\begin{tcolorbox}[title=\textbf{Prompt (P2) Example for Video Metadata Generation},
  colback=gray!3, colframe=black!40, boxrule=0.4pt, arc=2pt,
  left=2pt, right=2pt, top=2pt, bottom=2pt,
  width=\textwidth, breakable]

\begin{lstlisting}[style=promptstyle]

---
### Output Format (Strict JSON)

{
  "editing_quality": "high",
  "interviewee": "Cillian Murphy",
  "editing_justification": 
  "The editing uses smooth transitions, consistent pacing, cinematic overlay text, and strong color grading.",
  "suitable_for_dataset": "Yes",
  "suitability_justification": 
  "The video features widescreen cinematic movie clips integrated into the interview.",
  "timestamp_breakdown": [
    {
      "timestamp_start": "0:00",
      "timestamp_end": "0:04",
      "scene_description": "60 Minutes ticking stopwatch intro graphic.",
      "source_movie": null,
      "type": "Intro Graphic",
      "b_roll_purpose": null,
      "b_roll_uniqueness": null,
      "b_roll_uniqueness_justification": null,
      "source_search_queries": null
    },
    {
      "timestamp_start": "0:24",
      "timestamp_end": "0:31",
      "scene_description": "Cillian Murphy as Oppenheimer walking toward camera in Los Alamos.",
      "source_movie": "Oppenheimer",
      "type": "B-Roll (Movie Clip)",
      "b_roll_purpose": "Illustrates actor's performance in his most recent role.",
      "b_roll_uniqueness": "Unique",
      "b_roll_uniqueness_justification": 
      "This clip is directly tied to Murphy's portrayal in Oppenheimer.",
      "source_search_queries": [
        "Cillian Murphy Oppenheimer hat smoking",
        "Oppenheimer walking scene",
        "Cillian Murphy Los Alamos clip",
        "Oppenheimer intro scene",
        "Christopher Nolan Oppenheimer footage"
      ]
    }
  ]
}
\end{lstlisting}

\end{tcolorbox}
\vspace{-4pt}
\captionof{figure}{The prompt (P2) used for video analysis in \vebenc metadata generation. 
It asks the model to perform editing-quality evaluation, dataset suitability judgment, interviewee identification,
and timestamp-level scene annotation in strict JSON format.}
\label{fig:prompt_p2}
\end{minipage}
\end{center}
\twocolumn
\clearpage

\clearpage
\onecolumn
\begin{center}
\begin{minipage}{0.96\textwidth}
\begin{tcolorbox}[title=\textbf{Video Metadata Example},
  colback=gray!3, colframe=black!40, boxrule=0.4pt, arc=2pt,
  left=2pt, right=2pt, top=2pt, bottom=2pt,
  width=\textwidth, breakable]

\begin{lstlisting}[style=promptstyle]
"qj5MvD1bpMU": {
  "editing_quality": "high",
  "interviewee": "Timothée Chalamet",
  "editing_justification": "The video exemplifies high-quality broadcast journalism editing. It seamlessly integrates traditional A-Roll sit-down interviews, on-location walk-and-talk segments, and a vast array of B-Roll from a dozen different films and archival sources. The pacing is excellent, balancing introspective moments with dynamic musical performances. Transitions are smooth, and the use of film clips is always purposeful, directly illustrating the topics being discussed, such as Chalamet's acting process, his specific roles, or his personal history. The audio mix is clean, and the color grading is consistent across the interview segments, creating a cohesive and professional final product.",
  "suitable_for_dataset": "Yes",
  "suitability_justification": "This video is highly suitable. It is a professionally edited, widescreen (16:9) piece that features a significant amount of cinematic B-Roll from numerous high-profile movies. The edits are strong and meaningful, using the film clips to enhance the interview narrative. It contains no vertical or user-generated content.",
  "timestamp_breakdown": [
      ...
        {
          "timestamp_start": "0:51",
          "timestamp_end": "0:57",
          "scene_description": "The 60 Minutes stopwatch graphic appears again, indicating a commercial break.",
          "source_movie": null,
          "type": "Outro Graphic",
          "b_roll_purpose": null,
          "b_roll_uniqueness": null,
          "b_roll_uniqueness_justification": null,
          "source_search_queries": null
        },
        {
          "timestamp_start": "0:57",
          "timestamp_end": "1:23",
          "scene_description": "A-Roll of Timothée Chalamet being interviewed by Anderson Cooper. The setting is a dimly lit space that appears to be a music venue or studio, with a purple-lit background. Chalamet discusses his high level of commitment to his roles.",
          "source_movie": null,
          "type": "A-Roll (Interview)",
          "b_roll_purpose": null,
          "b_roll_uniqueness": null,
          "b_roll_uniqueness_justification": null,
          "source_search_queries": null
        },
        {
          "timestamp_start": "1:23",
          "timestamp_end": "1:50",
          "scene_description": "A montage of scenes featuring Timothée Chalamet as Bob Dylan. He is silhouetted against a spotlight, then seen from behind as he approaches a microphone on a stage in a large, classic theater, singing \"A Hard Rain's a-Gonna Fall\".",
          "source_movie": "A Complete Unknown",
          "type": "B-Roll (Movie Clip)",
          "b_roll_purpose": "To visualize the central topic of the interview: Chalamet's portrayal of Bob Dylan and the immense preparation involved.",
          "b_roll_uniqueness": "Highly unique",
          "b_roll_uniqueness_justification": "This is pre-release footage from a major biographical film. It is irreplaceable for discussing this specific role and showcases the actor's transformation, which is the core of the segment.",
          "source_search_queries": [
            "Timothée Chalamet A Complete Unknown trailer",
            "Timothée Chalamet Bob Dylan singing",
            "A Hard Rain's a-Gonna Fall A Complete Unknown",
            "James Mangold Bob Dylan movie clip",
            "Timothée Chalamet as Bob Dylan on stage"
          ]
        },
        ...
\end{lstlisting}

\end{tcolorbox}
\vspace{-4pt}
\captionof{figure}{Operation simulation metadata example.}
\label{fig:metadata_example}
\end{minipage}
\end{center}
\twocolumn
% \clearpage

\section{Open-Source Omni Model Evaluation}
\label{supp:omni}

To further examine the capability of open-source omni-directional multimodal models on real-world video editing tasks, we evaluate two representative Omni models on \vebenc, \textbf{VITA-1.5}~\cite{fu2025vita} and \textbf{VideoLLaMA3}~\cite{damonlpsg2025videollama3}. 
Both models are tested using the default frame-sampling strategy.
Table~\ref{tab:omni} reports the performance of the evaluated Omni models across all the subtasks. Despite their audio-visual processing capabilities, both models exhibit clear limitations on \vebenc.  
They achieve only moderate performance on TechRec and OpSim-FS, and similar to most open-source LMMs, they fail to complete the temporal localization component of OpSim, resulting in near-zero tIoU.  
These findings further underscore the gap between current open-source Omni models and the demands of realistic video editing workflows, which require nuanced temporal reasoning, multi-video understanding, and fine-grained multimodal perception.

\begin{table*}[t]
\centering
\caption{Evaluation results of omni models on \textbf{\vebenc}. ``TechRec'' denotes the \textit{Technique Recognition} subtask, and ``OpSim'' denotes the \textit{Operation Simulation} subtask, where ``FS'' represents \textit{Footage Selection}. The overall score is the average of TechRec and OpSim-FS. \textbf{Bold} numbers indicate the best performance.}
\resizebox{0.9\textwidth}{!}{
\begin{tabular}{@{}lcccc@{}}
\toprule
\textbf{Models} & 
\textbf{Overall (\%)} &
\textbf{TechRec (\%)} &
\textbf{OpSim-FS (\%)} &
\textbf{OpSim (\%)} \\
\midrule

VITA-1.5~\cite{fu2025vita} & 24.62 & 22.16 & 27.08 & 0.00 \\
VideoLLaMA3~\cite{damonlpsg2025videollama3} & 24.50 & 23.20 & 25.80 & 0.01 \\
% Qwen3-Omni-30B-A3B &  &  &  &  \\

\bottomrule
\end{tabular}
}
\label{tab:omni}
\vspace{-3pt}
\end{table*}

\section{OpSim-FS in Frame Index Evaluation Setting}
\label{supp:FI setting}
The near-zero scores on the full OpSim reflect a systematic failure in temporal grounding rather than an ill-posed evaluation, which aligns with VEBench’s goal of diagnosing the gap between \emph{what} content to use and \emph{how/where} to execute edits.
As shown in the Table~\ref{tab:opsim_frame_index}, frame-index (FI) evaluation brings limited improvement overall, which indicates that the difficulty of OpSim stems from intrinsic challenges in temporal decision-making, rather than the evaluation setting.

% \vspace{-10pt}
\begin{table*}[h]
\centering
\caption{Operation Simulation performance under frame-index-based evaluation.}
\small
\resizebox{\linewidth}{!}{
\begin{tabular}{lcccccc}
\toprule
\textbf{Model} &
Gemini-2.5-Pro &
GPT-4o &
Qwen3-VL-8B &
Qwen3-VL-4B &
Qwen2.5-VL-7B &
InternVL3-8B \\
\midrule
\textbf{FI tIoU} &
0.15 & 0.08 & 0.04 & 0.05 & 0.04 & 0.00 \\
\bottomrule
% \vspace{-50pt}
\end{tabular}
}
\label{tab:opsim_frame_index}
\end{table*}

\section{Data Imbalance Analysis}
\label{supp:data_imbalance}

This distribution largely reflects real-world editing practice, where techniques, such as Invisible Cut, are used much less frequently than more common transitions (J-Cut), and often appear only in specific genres or stylistic contexts (e.g., Invisible Cuts in one-shot illusion).
We provide a detailed accuracy breakdown by cut type here.
As shown in the table, performance remains consistently low not only for rare techniques (Invisible Cut), but also for frequently occurring ones (L-Cut).
This indicates that the observed performance degradation reflects the intrinsic difficulty of recognizing subtle editing semantics.

% \vspace{-10pt}
\begin{table*}[h]
\centering
\caption{Accuracy Breakdown of \textbf{TechRec} by cut type.}
\small
\setlength{\tabcolsep}{4pt}
\resizebox{\linewidth}{!}{
\begin{tabular}{lcccccccc}
\toprule
\textbf{Model} &
\textbf{J-Cut} &
\textbf{L-Cut} &
\textbf{Cutaway} &
\textbf{Match} &
\textbf{Jump} &
\textbf{Invisible} &
\textbf{Smash} &
\textbf{Overall} \\
\midrule
\textbf{InternVL3-8B} &
24.66 & 23.08 & 27.74 & 31.08 & 18.87 & 2.86 & 21.88 & 25.21 \\
\textbf{Gemini-2.5-Pro} &
38.54 & 31.24 & 38.99 & 35.29 & 36.36 & 13.64 & 29.73 & 34.65 \\
\textbf{GPT-4o} &
10.63 & 31.65 & 20.59 & 22.54 & 22.64 & 11.43 & 15.62 & 24.68 \\
\bottomrule
% \vspace{-30pt}
\end{tabular}
}
\label{tab:techrec_cuttype_horizontal}
\end{table*}

\section{More Visualizations}
\label{supp:visualization}
To further illustrate the difficulty and diversity of the \textbf{Operation Simulation} task in \vebenc, we provide two additional qualitative examples. Each visualization compares the predictions of three representative multimodal large models: Gemini-2.5-Pro~\cite{google2025gemini25pro}, Qwen3-VL-8B-Instruct~\cite{qwen3vl}, and InternVL3-8B~\cite{zhu2025internvl3}.
In both examples, the stitched video is composed of five temporally ordered segments: one reference video followed by four candidate option videos (A–D). The model must identify the \emph{only} option containing the most suitable continuation of the reference clip and localize the precise timestamps within that option segment. We show the full stitched timeline, annotated option descriptions, and each model’s predicted answer along with the ground truth and temporal IoU score.

\begin{figure*}[t]
  \centering
   \includegraphics[width=.95\linewidth]{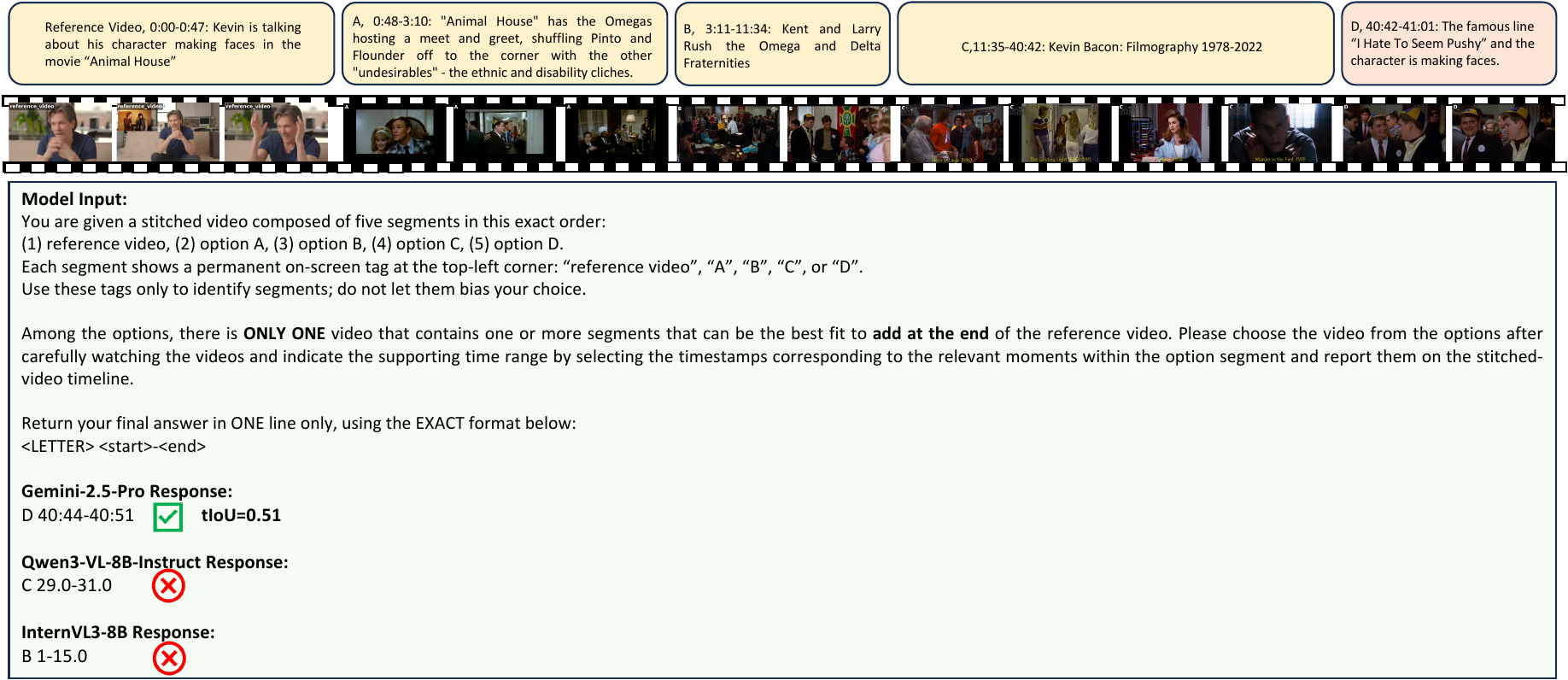}
   \caption{Qualitative example of Gemini-2.5-Pro~\cite{google2025gemini25pro}, Qwen3-VL-8B-Instruct~\cite{qwen3vl}, and InternVL3-8B~\cite{zhu2025internvl3}. }
   \label{fig:visualization2}
   \vspace{-3pt}
\end{figure*}

In the first example shown in Figure~\ref{fig:visualization2}, the reference video features Kevin Bacon describing how his character makes exaggerated facial expressions in the film Animal House. The task asks the model to identify which option contains the best segment to append at the end of this reference clip. Ideally, the correct answer should be a moment in Animal House where Kevin Bacon’s character is visibly making faces. Option A contains a related scene from the film but does not correspond to the specific moment described. Option B is thematically relevant but still does not show the correct facial-expression sequence. Option C contains a montage of Kevin Bacon’s representative works from 1978–2022, which is entirely mismatched for this context. In contrast, Option D includes the iconic “I hate to seem pushy” scene, where Bacon’s character performs the exact facial expression referenced in the interview. Gemini-2.5-Pro successfully identifies this segment with a high tIoU, while both Qwen3-VL-8B-Instruct and InternVL3-8B fail completely, selecting unrelated clips.

\begin{figure*}[t]
  \centering
   \includegraphics[width=.95\linewidth]{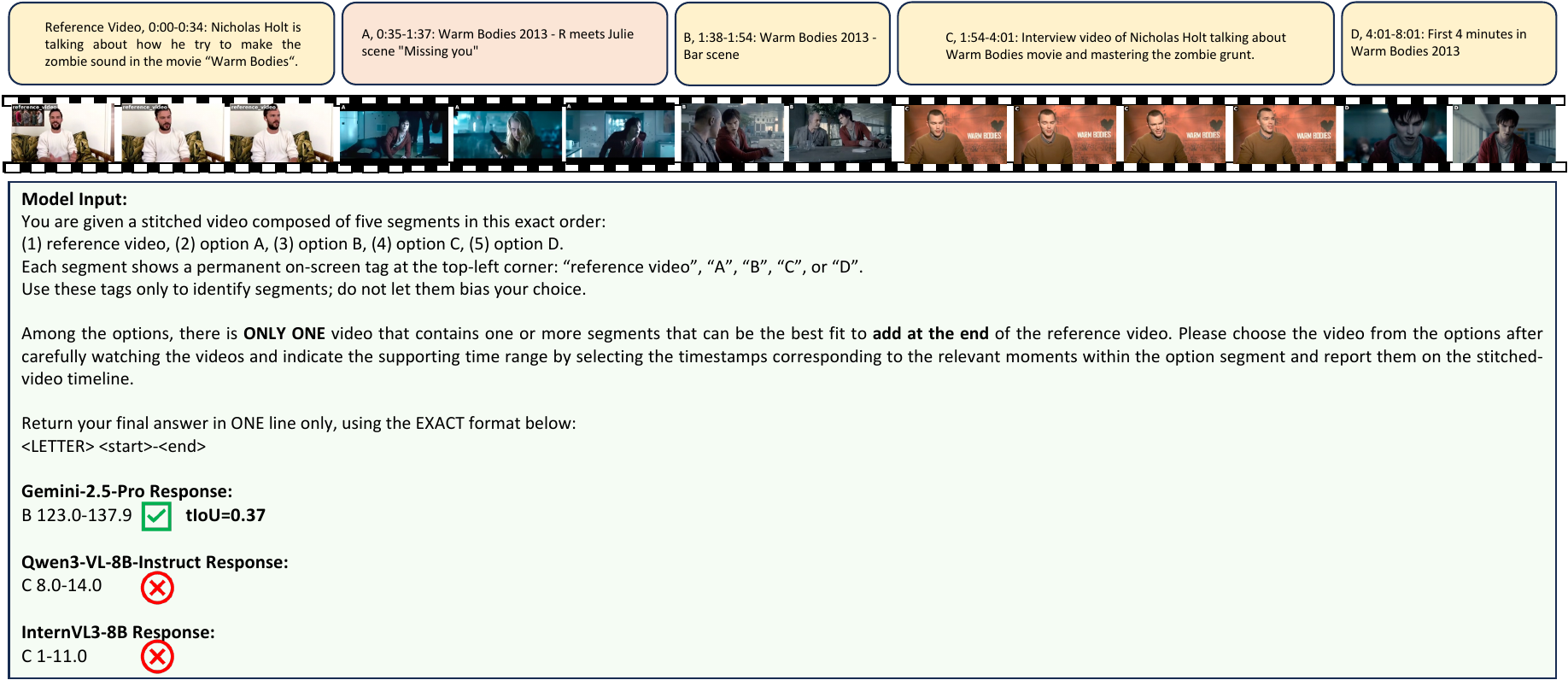}
   \caption{Qualitative example of Gemini-2.5-Pro~\cite{google2025gemini25pro}, Qwen3-VL-8B-Instruct~\cite{qwen3vl}, and InternVL3-8B~\cite{zhu2025internvl3}. }
   \label{fig:visualization3}
   \vspace{-3pt}
\end{figure*}

In the second example, shown in Figure~\ref{fig:visualization3}, the reference video contains an interview in which Nicholas Hoult explains how he created the zombie vocalization for his role in Warm Bodies. Therefore, when asked to identify the best segment to append at the end of the reference clip, the correct choice should be a scene from the film where Hoult’s character produces the distinctive zombie sound he describes. Among the four options, only Option B, the bar scene, includes a moment where the zombie emits a recognizable guttural sound toward another zombie. Gemini-2.5-Pro successfully captures this audio–visual cue and selects the correct segment with a strong tIoU. In contrast, the two open-source models fail entirely, largely due to their lack of audio-processing capability, preventing them from understanding the crucial auditory evidence required for this task.

These examples highlight the unique challenges posed by \vebenc, including cross-video semantic matching, fine-grained temporal localization, and reasoning about cinematic editing structure.

% \newpage
% {
%     \small
%     \bibliographystyle{ieeenat_fullname}
%     \bibliography{main}
% }

\end{document}